\documentclass[10pt,journal,compsoc]{IEEEtran}
\ifCLASSOPTIONcompsoc
  \usepackage[nocompress]{cite}
\else
  \usepackage{cite}
\fi
\ifCLASSINFOpdf
\else
\fi
\usepackage{bm}
\usepackage{amsmath}
\usepackage{amssymb}
\usepackage{subfigure}
\usepackage{amsmath}
\usepackage{diagbox}
\usepackage{adjustbox}
\usepackage{makecell}
\usepackage{mathrsfs}
\usepackage{xcolor}
\usepackage{balance}
\usepackage[pagebackref=true,breaklinks=true,letterpaper=true,colorlinks,bookmarks=false]{hyperref}

\usepackage[T1]{fontenc}

\newtheorem{thm}{Theorem}
\newtheorem{proposition}[thm]{Proposition}

\begin{document}
%
\title{MOPS-Net: A Matrix Optimization-Driven Network for Task-Oriented 3D Point Cloud Downsampling}

\author{Yue Qian, 
      Junhui Hou, ~\IEEEmembership{Senior Member,~IEEE,}, 
            Qijian Zhang, 
      Yiming Zeng, \\
      Sam Kwong, ~\IEEEmembership{Fellow,~IEEE,}
      and Ying He
      \IEEEcompsocitemizethanks{
       \IEEEcompsocthanksitem 
      This work was supported in part by the Hong Kong Research Grants Council under grants 9048123 and 9042820, and in part by the Natural Science Foundation of China under Grants 61871342. (\textit{Corresponding author: Junhui Hou})
      \IEEEcompsocthanksitem Y. Qian, J. Hou,  Q. Zhang, Y. Zeng, and S. Kwong are with the Department of Computer Science, City University of Hong Kong, Hong Kong. Email: yueqian4-c@my.cityu.edu.hk; jh.hou@cityu.edu.hk; ym.zeng@my.cityu.edu.hk; qijizhang3@cityu.edu.hk; cssamk@cityu.edu.hk
      \IEEEcompsocthanksitem Y. He is with the School of Computer Science and Engineering, Nanyang Technological University, Singapore, 639798. Email: yhe@ntu.edu.hk}}

\IEEEtitleabstractindextext{%
\begin{abstract}
This paper explores the problem of task-oriented downsampling over 3D point clouds, which aims to downsample a point cloud while maintaining the performance of subsequent applications applied to the downsampled sparse points as much as possible.
Designing from the perspective of matrix optimization, we propose \textit{MOPS-Net}, a novel interpretable deep learning-based method, which is fundamentally different from the existing deep learning-based methods due to its interpretable feature. The optimization problem is challenging due to its \textit{discrete} and \textit{combinatorial} nature. We tackle the challenges by relaxing the binary constraint of the variables,
and formulate a constrained and differentiable matrix optimization problem. 
We then design a deep neural network to mimic the matrix optimization by exploring both the local and global structures of the input data. MOPS-Net can be end-to-end trained with a task network and is permutation-invariant, making it robust to the input. We also extend MOPS-Net such that a single network after one-time training is capable of handling arbitrary downsampling ratios.
Extensive experimental results show that MOPS-Net can achieve favorable performance against state-of-the-art deep learning-based methods over various tasks, including classification, reconstruction, and registration. Besides, we validate the robustness of MOPS-Net on noisy data. 
\end{abstract}

}

\maketitle

\IEEEdisplaynontitleabstractindextext

%
\IEEEpeerreviewmaketitle

\IEEEraisesectionheading{\section{Introduction}\label{sec:introduction}}

%
%
%
%
\IEEEPARstart{W}{ith} recent advances in three-dimensional (3D) sensing technology (e.g., LiDAR scanning devices), 3D point clouds can be easily obtained. Compared with other 3D representations such as multi-view images, voxel grids and polygonal meshes, point clouds are a raw 3D representation, containing only 3D samples which are located on the scanned surface. Powered by deep learning techniques, the performance of many point cloud applications, such as classification, segmentation and reconstruction, has been improved significantly in recent years. However, processing large-scale and/or dense 3D point clouds is still challenging due to the high cost of computation, storage, and communication load.

\begin{figure}[t]
\centering
\includegraphics[width=0.5\textwidth]{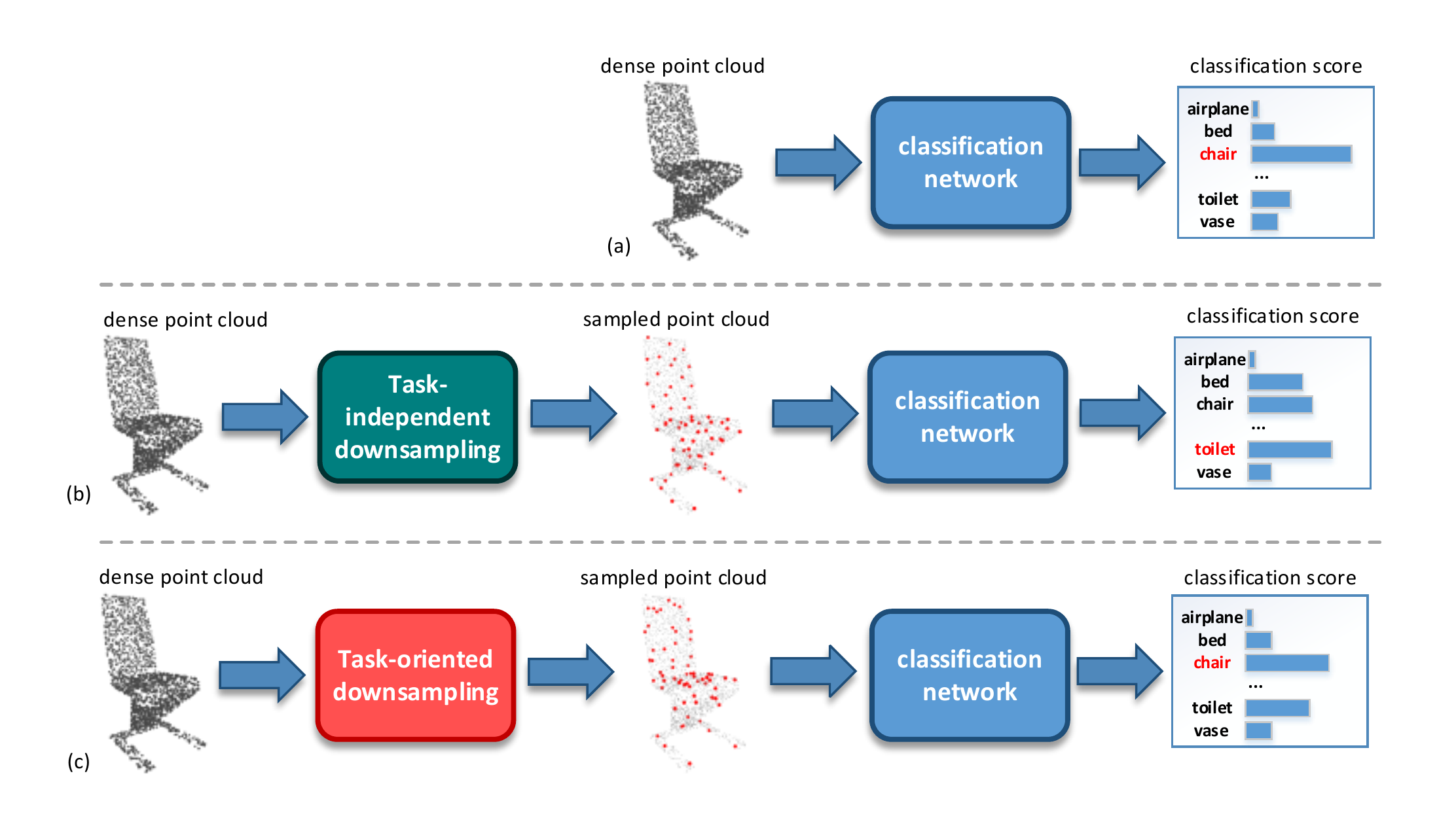}\vspace{-0.4cm}
\caption{Illustrations of task-independent and task-oriented point cloud downsampling using classification as an example. (a) A deep learning based point cloud classifier is trained on dense point clouds. (b) Classic downsampling methods, such as FPS, generate sparse point clouds without considering the nature of the task, hereby may compromise the performance of the classifier in (a) significantly. (c) The classification-oriented downsampling method produces sparse point clouds that maintain the performance of the classifier in (a). To develop such a classification-oriented downsampling, we take both the geometry of the input shape and the performance of the classifier into account.  
}
\label{fig:illustration of the problem}
\end{figure}

Point cloud downsampling is a popular and effective technique to reduce information redundancy, hereby improving the runtime performance of the downstream applications and saving storage space and transmission bandwidth. The classic downsampling approaches such as farthest point sampling (FPS)~\cite{fps} and Poisson disk sampling (PDS)~\cite{DBLP:journals/tvcg/YingXS013} iteratively generate uniformly distributed samples on the input shape, and thus they can preserve the geometry well. Such sampling methods, however, focus on reducing the geometry loss only and are completely independent of the downstream applications. Although voxelization is computationally efficient, it suffers from quantization errors. As a result, the downsampled point clouds yielded by the classic methods would degrade the performance of the subsequent applications severely.

An alternative way for downsampling is to generate samples that optimize the performance of a particular task, i.e., the resulting sparse point clouds will maintain the task performance as much as possible. Due to the task centric nature, we call it \textit{task-oriented downsampling}.  Moreover, an effective downsampling method should allow the user to freely specify the downsampling ratio to balance the task performance and computation efficiency.

As the deep learning technologies have proven effective in point cloud classification~\cite{qi2017pointnet++,qi2017pointnet,li2018pointcnn,shen2018mining,wang2019dynamic} and segmentation~\cite{li2018so,tchapmi2017segcloud,su2018splatnet,hu2020randla}, it is highly desired to combine downsampling methods with the deep neural networks. 
However, extending the existing network architectures to point cloud downsampling is non-trivial, since point selection process is discrete and non-differentiable. Recently, Dovrat et. al. pioneered a deep learning approach, called S-Net~\cite{dovrat2019learning}, that takes downsampling as a generative task and uses the extracted global features to generate samples. S-Net is flexible in that it can be combined with task-specific networks to produce an end-to-end network trained by a joint loss. Thanks to its task-oriented nature, S-Net outperforms FPS in various applications. Hereafter, Lang \textit{et al.} ~\cite{lang2020samplenet} improved S-Net by introducing  additional projection module to encourage the generated points closer to original point clouds, learning SampleNet. However, as S-Net and SampleNet solely relies on global features in its generative process, they can not  utilize the point-wise high-dimensional local features, which limits the quality of synthesized points.

In this paper, we propose a novel deep learning approach, called MOPS-Net, for task-oriented point cloud downsampling. In contrast to the existing methods, we propose MOPS-Net from the matrix optimization perspective. Viewing downsampling as a selection process, we first formulate a discrete and combinatorial optimization problem. As it is difficult to solve the 0-1 integer program directly, we relax the integer constraint for each variable, in which a constrained and differentiable matrix optimization problem with an implicit objective function is formulated. 
We then design a deep neural network architecture to mimic the matrix optimization problem by exploring both the local and the global structures of the input data. 
With a task network, MOPS-Net can be end-to-end trained. MOPS-Net is permutation-invariant, making it robust to input data. Moreover,  by restricting the invoking columns of the soft sampling matrix, we also extend MOPS-Net such that a single network with one-time training is capable of handling arbitrary downsampling ratios. Extensive results show that MOPS-Net achieves much better performance than state-of-the-art deep learning-based methods and  traditional methods in various tasks, including point cloud classification, reconstruction, and registration. We also provide comprehensive ablation studies and analysis on model robustness and time efficiency.

The main contributions of this paper are summarized as follows:\vspace{-0.25cm}
\begin{itemize}
    \item  we explicitly formulate the problem of task-oriented point cloud downsampling from the matrix optimization perspective;
    \item  we propose an interpretable deep learning-based method named MOPS-Net, which mimics the above formulation, making it fundamentally different from the existing deep learning-based methods;
    \item we extend MOPS-Net and propose FMOPS-Net, which is capable of handling arbitrary downsampling ratios after only one-time training; and 
    \item we propose a new compact deep-learning framework for large-scale point cloud reconstruction, which enables the validation of the scalability of MOPS-Net on relatively large-scale point clouds; and 
    \item we conduct various experiments and comprehensive ablation studies to demonstrate the advantages of our methods over state-of-the-art methods.
\end{itemize}

The rest of this paper is organized as follows. Section \ref{sec:RW} reviews classic point cloud downsampling methods and recent deep learning techniques for 3D point clouds. In Section \ref{sec:formulation}, we formulate task-oriented downsampling as a constrained and differentiable optimization problem, followed by an end-to-end task-oriented downsampling deep neural network which mimics the resulting optimization in Section \ref{sec:proposed}. Section \ref{sec:exp} presents extensive experimental results, comparisons with the state-of-the-art, as well as comprehensive robustness tests and ablation studies. Section \ref{sec:con} finally concludes this paper.

\begin{figure*}[htp!]
\centering
\includegraphics[width=1\textwidth]{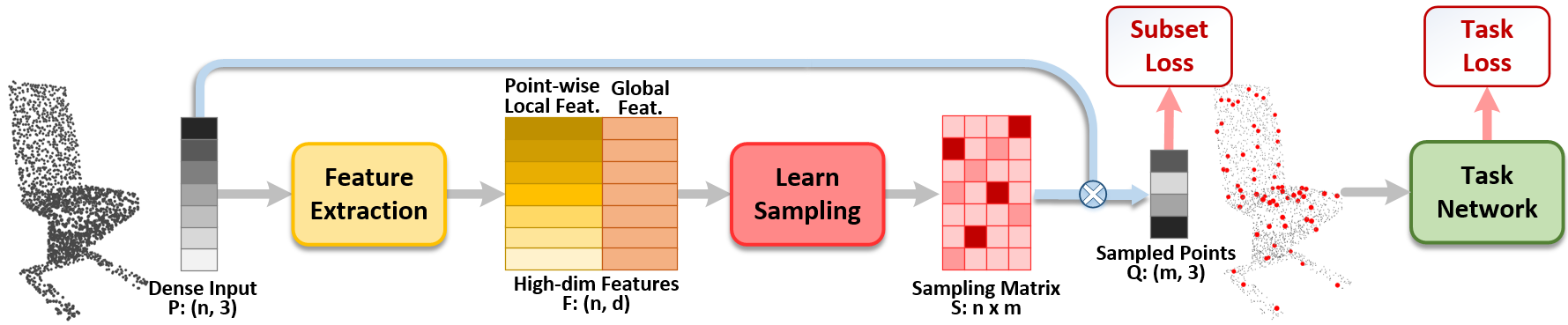}
\caption{MOPS-Net is a matrix optimization driven deep learning method for task-oriented 3D point cloud downsampling. It first extracts high-dimensional features,  
which contain both global and local geometric information, from point coordinates (Sec. \ref{subsec:method_fea}).
It then utilizes the features to learn a differentiable sampling matrix, which is multiplied to the input dense point cloud to obtain the sampled points (Sec \ref{subsec:method_S}). Finally, it feeds the downsampled sparse points to a task network. The whole network is trained by jointly 
optimizing the task loss and the subset loss (Sec. \ref{subsec:loss}).}
\label{fig:architecture}
\end{figure*}

\section{Related work}
\label{sec:RW}

\subsection{Traditional Downsampling Methods}
The traditional methods, such as farthest point sampling (FPS) and Poisson disk sampling (PDS), generate samples in an iterative manner. Starting from a random sample, FPS repeatedly places the next sample point in the center of the least-sampled area. Using efficient geodesic distance computation tools (such as the fast marching method~\cite{Kimmel8431}), FPS generates $m$ samples on a $n$-vertex mesh in $\sum_{i=1}^{m}f(\frac{n}{i})=O(n\log n)$ time, where $f(x)=O(x\log x)$. 
FPS is easy to implement and becomes popular in designing neural networks that aggregate local features \cite{qi2017pointnet++,li2018pointcnn,wu2019pointconv}. Poisson disk sampling produces samples that are tightly-packed, but no closer to each other than a specified minimum distance, resulting in a more uniform distribution than FPS. There are efficient implementations of PDS with linear time complexity in Euclidean spaces~\cite{bridson2007fast,wei2008parallel}. However, it is expensive to generate Poisson disk samples on curved surfaces due to frequent computation of geodesic distances~\cite{DBLP:journals/tvcg/YingXS013}.  Voxelization is also a commonly used technique to downsample or resample point clouds, which quantizes point clouds into regular voxels in 3D space with predefined resolution. Compared with FPS and PDS, voxelization is more efficient due to its non-iterative manner. However, voxelization often suffers from quantization error and cannot yield results that are the exact subsets of the input. Moreover, the traditional approaches focus only on preserving the geometry of the input shape and they do not consider the downstream tasks at all. 

\subsection{Deep Learning for 3D Point Clouds}  
Due to the irregular and unordered nature of point clouds, the widely used convolutional neural networks (CNNs) on 2D images/videos~\cite{rastegari2016xnor,krizhevsky2012imagenet,he2016deep} cannot be applied directly. PointNet,  proposed by Qi \textit{et al.}~\cite{qi2017pointnet}, maps a 3D point to  a high dimensional space by point-wise multi-layer perceptrons (MLPs) and aggregates global features by a symmetric function, named max-pooling. As the first deep neural network that works for 3D raw points without projecting or parameterizing them to regular domains, PointNet quickly gained popularity and was successfully used as the fundamental feature extraction for point clouds.  However, PointNet processes the points individually and does not consider the spatial relation among points. The follow-up works, such as PointNet++ \cite{qi2017pointnet++}, DGCNN \cite{wang2019dynamic} and PointCNN \cite{li2018pointcnn}, improve PointNet by taking local geometry into account. 

Inspired by the success of PointNet on classification, many other point cloud applications were studied in recent years, such as retrieval \cite{angelina2018pointnetvlad,kuang2018deep}, segmentation \cite{li2018so,tchapmi2017segcloud,su2018splatnet,hu2020randla}, reconstruction \cite{achlioptas2018learning,sun2020pointgrow,li2018point}, registration \cite{elbaz20173d,aoki2019pointnetlk,yew20183dfeat,zaganidis2018integrating}, and object detection \cite{zhou2018voxelnet,li20173d,shi2019pointrcnn,engelcke2017vote3deep,lang2019pointpillars}, just name a few. Although they have different problem settings, these networks can be combined with the point cloud downsampling network and jointly trained. In this paper, we evaluate the performance of the proposed downsampling framework on classification, registration, and reconstruction. 

Opposite to downsampling, point cloud upsampling \cite{yu2018pu,yifan2019patch,li2019pu,qian2020pugeo} has also been investigated recently. Upsampling can be treated as either a 3D version of image super-resolution, or the inverse process of downsampling. Despite the common word ``sampling'', the two tasks are completely different. Upsampling, as a generative task, requires informative feature expansion and can be trained by ground truth dense point clouds. In contrast, point cloud downsampling is close to feature selection, where a differentiable end-to-end framework should be carefully designed. Moreover, due to lack of ground truth, the downsampled points should be learned to optimize a specific task loss.

\subsection{Deep Learning-based Point Cloud Downsampling}

It is an emerging topic, on which there are only a few works. Nezhadary \textit{et al.}~ \cite{nezhadarya2020adaptive} proposed to use critical points invoked in max-pooling as sampled points. In order to improve classification accuracy, Yang \cite{yang2019modeling} adopted a gumbel softmax layer to integrate high level features. Recently, Dovrat \textit{et al.} \cite{dovrat2019learning} proposed a data-driven point cloud downsampling framework named S-Net. After the point-wise feature extraction by PointNet, a global feature was obtained by the max-pooling operation. Then 3D coordinates of fewer points were regressed by fully-connected layers. Followed by a pre-trained task network, S-Net can be trained end-to-end.  Lang \textit{et al.}~\cite{lang2020samplenet} proposed SampleNet, which improves S-Net by introducing a soft projection module that adopts an anealing schedule to encourage generated points to be close to original points. However, both S-Net and SampleNet regress coordinates from global features directly and do not consider spatial correlation among sampled points, which plays an important role in downsampling, since spatially close points have the tendency to be represented by the same downsampled point.  In sharp contrast to S-Net and SampleNet, we propose a novel framework by exploring the local geometry of the input data from a perspective of matrix optimization.

\section{Problem Formulation}
\label{sec:formulation}
In this section, we explicitly formulate the problem of task-oriented point cloud downsampling from the matrix optimization perspective.

Denote by $\mathcal{P}=\{\mathbf{p}_i\in \mathbb{R}^3\}_{i=1}^n$ a dense point cloud with $n$ points. $\mathbf{p}_i=\{x_i,y_i,z_i\}$ is the 3D Cartesian coordinates. Let $\mathcal{Q}=\{\mathbf{q}_i\in \mathbb{R}^3\}_{i=1}^m$ be the downsampled sparse point cloud with $m~(<n)$ points, which is a subset of $\mathcal{P}$, i.e., $\mathcal{Q}\subset \mathcal{P}$.

As aforementioned, we consider task-oriented downsampling. That is, given $\mathcal{P}$, we compute the downsampled point cloud $\mathcal{Q}$ that maintains comparable or at least does not compromise the performance of the subsequent tasks too much, e.g., classification, etc. Such a downsampling process is beneficial to computation, storage and transmission.

Mathematically speaking, the problem can be formulated as
\begin{equation}
    \min_{\mathcal{Q}} L_{task}(\mathcal{Q})  ~~s.t.~~ \mathcal{Q}\subset \mathcal{P}, 
    \label{eqn:origial}
\end{equation}
where $L_{task}(\cdot)$ indicates the task-tailored loss, whose explicit form will be discussed in Section \ref{subsec:loss}.  We rewrite Eq. (\ref{eqn:origial}) in a more explicit manner 
\begin{align}
    & \min_{\mathbf{S}} L_{task}(\mathbf{Q}) \nonumber \\ 
    &s.t.~\mathbf{Q}=\mathbf{S}^\textsf{T}\mathbf{P}, 
       ~\mathbf{1}_n^{\textsf{T}}\mathbf{S}=\mathbf{1}_m^{\textsf{T}}, ~\mathbf{S}^\textsf{T}\mathbf{S}=\mathbf{I}_m, ~s_{i,j}\in\{0, 1\},
    \label{eqn:discrete}
\end{align}
where $\mathbf{P}=[\mathbf{p}_1,\mathbf{p}_2,\dots,\mathbf{p}_n]^\textsf{T}\in\mathbb{R}^{n\times 3}$ and $\mathbf{Q}=[\mathbf{q}_1,\mathbf{q}_2,\dots,\mathbf{q}_m]^\textsf{T} \in\mathbb{R}^{m\times 3}$ are the matrix representations of $\mathcal{P}$ and $\mathcal{Q}$, respectively, constructed by stacking each point as a column in an \textit{unodered} manner\footnote{Note $\mathcal{P}$ (resp. $\mathcal{Q}$) and $\mathbf{P}$ (resp. $\mathbf{Q}$) stand for the same data but in different forms, and there is no specific requirement on the order when stacking the points. The notations are used interchangeably in the paper.}; 
$s_{i,j}$ is the $(i, j)$-th entry of $\mathbf{S}\in \mathbb{R}^{n\times m}$;  $\mathbf{1}_n=[1,\dots,1]^\textsf{T}\in\mathbb{R}^{n\times 1}$ is the column vector with all elements equal to 1; and $\mathbf{I}_m$ refers to the identity matrix of size $m\times m$. The constraints force $\mathbf{S}$ to be an ideal sampling matrix, i.e., $\mathbf{S}$ only contains $m$ columns of a permutation matrix of size $n\times n$. More precisely, the orthogonal constraint is used to avoid repeated columns (indicating that an identical point is selected multiple times) in $\mathbf{S}$.

The challenge for solving Eq. (\ref{eqn:discrete}) comes from the discrete and binary characteristics of matrix $\mathbf{S}$. To tackle this challenge, we relax the binary constraints Eq. (\ref{eqn:discrete}) in a soft and continuous manner, i.e., the elements of $\mathbf{S}$ are continuous, ranging between 0 and 1. The relaxed variables indicate the probabilities of the corresponding points that will be sampled. After this relaxation, the resulting points in $\mathcal{Q}$ may not be the subset of  $\mathcal{P}$. To mitigate this effect for a meaningful sampling process, we further introduce a metric $L_{subset}(\cdot,\cdot)$ to quantitatively measure the distant between two point clouds, and minimizing $L_{subset}(\mathbf{P},\mathbf{Q})$ will promote $\mathbf{Q}$ to be a subset of $\mathbf{P}$ as much as possible. We will explain the explicit form of $L_{subset}(\cdot, \cdot)$ in Section \ref{subsec:loss}.

 We finally express the \textit{relaxed} and \textit{ continuous} optimization problem for task-oriented point cloud downsampling as 
\begin{align}
    &\min_{\mathbf{S}} L_{task}(\mathbf{Q}) +\alpha L_{subset}(\mathbf{P}, \mathbf{Q})\nonumber \\
   & s.t.~\mathbf{Q}=\mathbf{S}^\textsf{T}\mathbf{P},~
         \mathbf{S}  \geq 0, ~
        \mathbf{1}_n^\textsf{T}\mathbf{S}  = \mathbf{1}_m^\textsf{T},~ \|\mathbf{S}^\textsf{T}\mathbf{S}-\mathbf{I}_m\|_F<\epsilon,  
\label{eqn:relax}
\end{align}
where $\|\cdot\|_F$ is the Frobenious norm of a matrix, $\epsilon>0$ is a threshold, and $\alpha>0$ is the penalty parameter to balance the two terms. 

\textit{Remarks}. Relaxing the binary matrix $\bf S$ produces a point cloud which is not a subset of the input $\mathcal{Q}\not\subset\mathcal{P}$. This can be explained from the perspective of geometry processing. When presenting an object using two point clouds of different resolutions, the one with fewer points is generally not fully overlapping with the larger one since we have to re-distribute the points in order to preserve the geometry. Although $\mathcal{Q}\not\subset\mathcal{P}$, the objective function penalizes the points that are away from the input shape. Therefore, the points of $\mathcal{Q}$ are either on or close to the underlying object surface.
If the subsequent task requires $\mathcal{Q}\subset\mathcal{P}$, we can assign each point of $\mathcal{Q}$ the closest point in the input data.  In Section \ref{sec:exp}, we will quantitatively analyze the effect of these post-processing operations. 


\section{Proposed Method}
\label{sec:proposed}
 
This section presents MOPS-Net, a novel end-to-end deep neural network that mimics the formulated optimization problem in Eq. (\ref{eqn:relax}) for task-oriented point cloud downsampling. 
\subsection{Overview}  
Figure~\ref{fig:architecture} illustrates the flowchart of MOPS-Net. Given a  point cloud $\mathcal{P}$ and a pre-trained task network, MOPS-Net uses a feature extraction module
to encode each point with high dimensional and informative features by exploring both the local and the global structures of $\mathcal{P}$ (Section \ref{subsec:method_fea}). Based on the high dimensional features, MOPS-Net  estimates a differentiable sampling matrix $\mathbf{S}$ under the guidance of the constraints
in Eq. (\ref{eqn:relax}).
Multiplying the 
learned sampling matrix to original dense point clouds, we can obtain the sampled points. 
Together with a fixed task network, MOPS-Net can be end-to-end trained with a joint loss (Section \ref{subsec:loss}). Such a joint loss simultaneously penalizes the degradation of task performance and regularizes the distribution of sampled points, which is consistent with the objective function in Eq. (\ref{eqn:relax}). We show that MOPS-Net is flexible and it can be extended so that a \textit{single} network with \textit{one-time} training is capable of handling \textit{arbitrary} downsampling ratio (Section \ref{subsec:FMOPS-Net}). Last but not the least, we prove that MOPS-Net is \textit{permutation-invariant}, which is a highly desired feature for point cloud applications. 

\textit{Remarks}. The proposed MOPS-Net is fundamentally different from the existing deep learning framework S-Net~\cite{dovrat2019learning} and SampleNet~\cite{lang2020samplenet}. They formulate the sampling process as a point generation problem from global features, while MOPS-Net is designed from matrix optimization perspective to utilize the informative local (or point-wise) features. Experimental results demonstrate the advantages of MOPS-Net over S-Net and SampleNet on point cloud classification, reconstruction and registration. See Section \ref{sec:exp}.

\subsection{Feature Extraction}
\label{subsec:method_fea}
Given a point cloud  $\mathcal{P}=\{\mathbf{p}_i\in\mathbb{R}^3\}_{i=1}^n$, we extract $d$-dimensional point-wise features, denoted by $\mathcal{F}=\{\mathbf{f}_i\in\mathbb{R}^d\}_{i=1}^n$. Let $\mathbf{F}=[\mathbf{f}_1,\dots, \mathbf{f}_n]^\textsf{T}\in\mathbb{R}^{n\times d}$ denote the matrix form of pointwise features. We utilize  PointNet~\cite{qi2017pointnet}, a basic feature extraction backbone over 3D point clouds,  to extract pointwise local features via a shared MLP, and a permutation-invariant global feature can be derived by applying the max-pooling operation to the resulting point-wise features along the spatial dimension. 
extract permutation-invariant features. 
We finally concatenate the global feature to each pointwise local feature to form $\mathbf{F}$.


It is also worth noting that other advanced feature extraction techniques, such as PointNet++~\cite{qi2017pointnet++}, DGCNN~\cite{wang2019dynamic}, and KPConv~\cite{thomas2019kpconv} , could be adopted to further boost the performance of our method. , Here we adopt the basic PointNet for fair comparisons with S-Net and SampleNet.  

\subsection{Learning Differentiable Sampling Matrix}
\label{subsec:method_S}

\begin{figure}[t]
\centering
\includegraphics[width=0.45\textwidth]{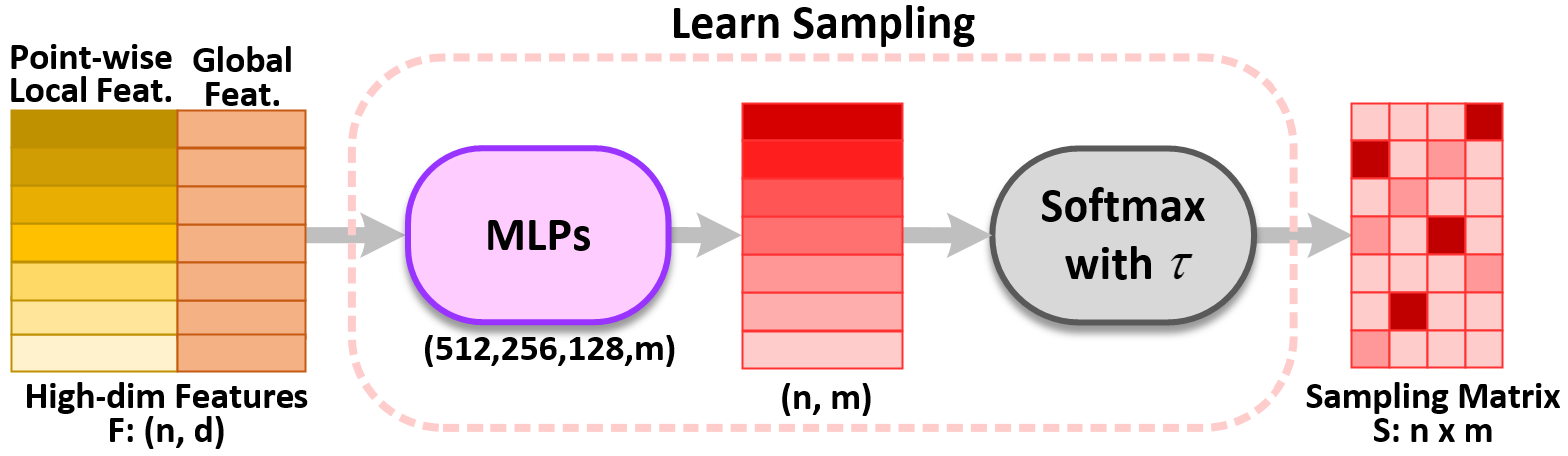}
\caption{The flowchart of differentiable sampling module.}
\label{fig:flowchart_sample}
\end{figure}

As analyzed in Section \ref{sec:formulation}, an ideal sampling matrix, which is a submatrix of a permutation matrix, is discrete and non-differentiable, making it challenging to optimize.  Accordingly, such a non-differentiable sampling matrix cannot be implemented in a deep neural network. Fortunately, the relaxation on the sampling matrix in Eq. (\ref{eqn:relax}) leads to a continuous matrix with additional constraints, which approximates the ideal one and allows us to design a deep neural network.

\begin{figure}[htp]
     \centering\vspace{-0.3cm}
        \subfigure[]{\includegraphics[width=1.7in]{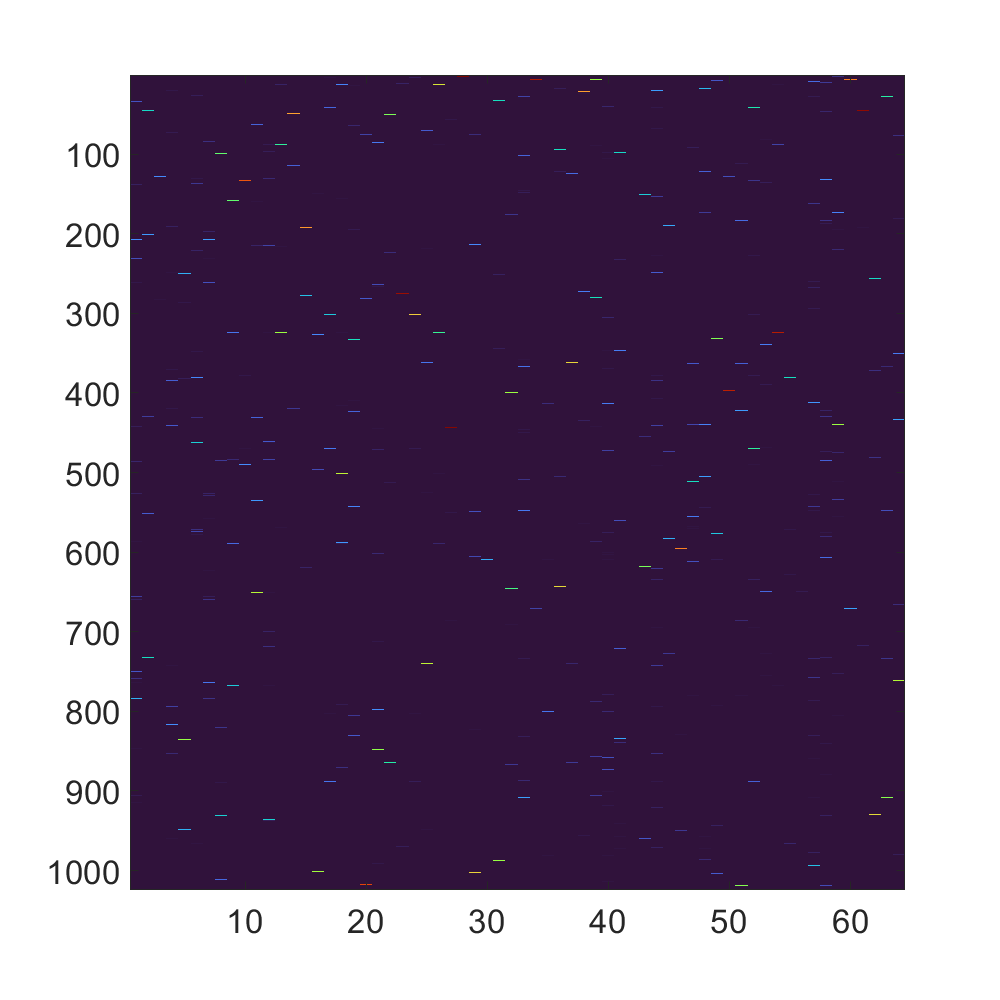}}
         \subfigure[]{\includegraphics[width=1.7in]{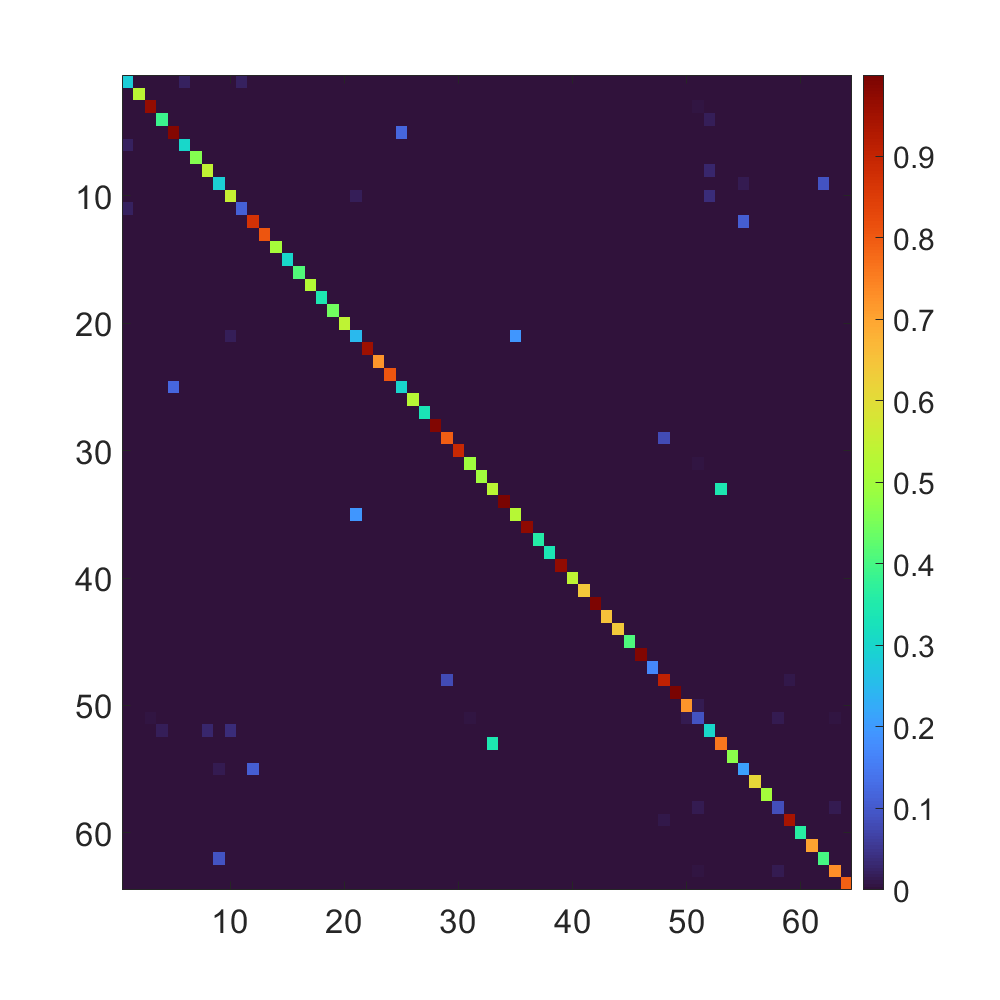}}\vspace{-0.4cm}
        \caption{Visualization of (a) learned sampling matrix $\mathbf{S}$ and (b) $\mathbf{S}^\textsf{T}\mathbf{S}$ for $m=64$. The two subfigures share the same colorbar.}
        \label{fig:sampling}
\end{figure}

According to Eq. (\ref{eqn:relax}), it is known that the optimized sampling matrix $\mathbf{S}$ will depend on $\mathcal{P}$ in a non-linear fashion.  From the perspective of geometry processing, downsampling highly depends on the geometric structure of the input data. 
As the high-dimensional embeddings $\mathbf{f}_i$ produced by the feature extraction module already encode such a structure locally and globally, we use them to predict a preliminary sampling matrix $\overline{\mathbf{S}}\in\mathbb{R}^{n\times m}$ row by row via an MLP $\rho(\cdot)$, i.e., 
\begin{equation}
\mathbf{\overline{s}}_i = \rho(\mathbf{f}_i),
\end{equation}
where $\overline{\mathbf{s}}_i\in\mathbb{R}^{1\times m}$ is the $i$-th row of $\overline{\mathbf{S}}$. In our experiments, $\rho(\cdot)$ is realized by a 4-layer MLP of size $[512, ~256,~128,~$m$]$. To satisfy the constraints on the sampling matrix in Eq.~(\ref{eqn:relax}), we further apply the  
softmax with temperature annealing operation on each element $\overline{s}_{ij}$ of $\mathbf{\overline{S}}$:  
\begin{equation}
s_{ij}=\frac{e^{\overline{s}_{ij}/\tau}}{\sum_{i=1}^n e^{\overline{s}_{ij}/\tau}},
\label{equ:annealing}
\end{equation}
where $s_{ij}$ and is the $(i, j)$-th entry of the final sampling matrix $\mathbf{S}$, and the value of the temperature $\tau$ gradually decreases from 1 to $\tau_{min}$ during training, and it is fixed to $\tau_{min}$ during inference.
Such an operation ensures $s_{ij}$ to be non-negative and encourages each column of $\mathbf{S}$ to be dominated by a single element, especially when $\tau$ is small. We experimentally found that such an operation is able to realize the constraints in Eq.~(\ref{eqn:relax}) well. As visualized in Figure~\ref{fig:sampling}, the learned sampling matrix $\mathbf{S}$ is extremely sparse and close to a sample matrix. Besides, $\mathbf{S}^\textsf{T}\mathbf{S}$ is also close to an identity matrix, although we do not explicitly minimize this constraint.

\subsection{Efficient Regression of the Sampled Set}
\label{subsec:method_sampledP}
Having obtained the sampling matrix $\mathbf{S}$, we can naturally deduce the corresponding downsampled point set $\mathbf{Q}$ by

\begin{equation}
\mathbf{Q}=\mathbf{S}^\textsf{T}\mathbf{P}.
\label{equ:regressed_point}
\end{equation}

Analytically, the downsampling procedure described in Eq.~(\ref{equ:regressed_point}) requires explicitly storing the sampling matrix $\mathbf{S}\in\mathbb{R}^{n\times m}$ and performing dot-product with $\mathbf{P}$ in $\mathcal{O}(nm)$ time complexity, which seems to be a major computational bottleneck faced with large-scale point clouds. Fortunately, benefiting from the structural sparsity of $\mathbf{S}$, we can perform the matrix multiplication operation in a much more efficient manner during inference.

More specifically, driven by the temperature annealing operation in Eq. (\ref{equ:annealing}), the learned sampling matrix $\mathbf{S}$ 
highly approximates 
an ideal sampling matrix, i.e., a column-truncated binary permutation matrix with only $m$ non-zero entries, and thus most entries of $\mathbf{S}$ are \textit{extremely} small (see Figure \ref{fig:sampling}) and make negligible contributions to the actual subset selection. In practice, we design a deterministic matrix simplification rule by setting all entries of $\mathbf{S}$ smaller than an appropriate threshold to zero, after which we can expect that the ratio of non-zero entries should be as small as $\frac{1}{n}$. Particularly, 
with the threshold set to $0.01$ in our implementation,  we counted 
the percentage of non-zero elements in $\mathbf{S}$. 
As listed in Table~\ref{table:nonzero},  we can observe that the number of non-zero elements of the quantized 
$\mathbf{S}$ is about $cm$, where $c$ is around 4. 
In our implementation, we adopt the Coordinate format (COO)~\cite{saad2003iterative} to store the highly sparse matrix $\bf S$, which has $\mathcal{O}(cm)$ memory cost, where $c \ll n$.
Therefore, performing dot-product on only non-zero elements has time complexity $\mathcal{O}(cm)$. 

As relaxation cannot guarantee $\mathcal{Q}\subset\mathcal{P}$, we can adopt an optional post-matching operation that maps each point of $\mathcal{Q}$ to its nearest point in $\mathcal{P}$ to obtain the downsampled subset. If the number of points contained in the matched point set is smaller than the specified value, we adopt FPS to complete it, i.e., points with farthest distances to the matched point set are iteratively selected from the original dense point clouds, until the target number is achieved. In Section \ref{sec:exp}, we will demonstrate the performance of the three cases.

\begin{table}[t]
\centering
\caption{Statistic of the percentage of non-zero elements in learned sampling matrix $\mathbf{S}$ ($n=1024$). $r$ is the threshold.}\vspace{-0.1in}
\begin{tabular}{c||ccc}\Xhline{5\arrayrulewidth}
$r$ & $m=32$ &  $m=64$ &  $m=128$  \\\Xhline{2\arrayrulewidth}
0.01& 0.48\% & 0.36\% & 0.33\% \\
\Xhline{5\arrayrulewidth}
\end{tabular}
\label{table:nonzero}
\end{table}

In addition to differentiability, our design of the sampling matrix learning is also permutation-invariant. That is, the sampling result is independent of the point order of the input data. 


\begin{table*}[t]
\centering
\caption{Comparisons of the classification accuracy by different downsampling methods. 
The larger, the better.}
\vspace{-0.1in}
\begin{tabular}{c||c|c|c|ccc|ccc|ccc|ccc}\Xhline{5\arrayrulewidth}
   &\makebox[2.6em]{RS}&\makebox[2.6em]{Voxel}&\makebox[3em]{FPS} &&\makebox[2.6em]{S-Net \cite{dovrat2019learning}}&&&\makebox[2.6em]{SampleNet  \cite{lang2020samplenet}}&&&\makebox[2.6em]{MOPS-Net}& &&\makebox[2.6em]{FMOPS-Net} &\\
      $m$ &  & & &\makebox[2.6em]{G}&\makebox[2.6em]{M}&\makebox[2.6em]{C}&\makebox[2.6em]{G}&\makebox[2.6em]{M}&\makebox[2.6em]{C}&\makebox[2.6em]{G}&\makebox[2.6em]{M}&\makebox[2.6em]{C} &\makebox[2.6em]{G}&\makebox[2.6em]{M}&\makebox[3em]{C}     \\\Xhline{2\arrayrulewidth}
 512 & 84.76 & 73.82 &  86.06 & 81.69 & 71.43& 85.66 & 57.82 & 57.82 & 86.63 & 86.67 & 85.25 & \bf{86.75}& 86.18& 85.07&  86.35\\
256 & 76.94& 73.50&   83.06&  82.94&  72.08&  82.78&  83.23&  83.02&  84.48&  \bf{86.63}& 85.53&  86.10& 86.51& 85.17&  86.14\\
128&62.32 &68.15  &72.85& 83.31&  72.24&  73.18&  84.04&  83.14&  82.58&  86.06&  \bf{84.64}& 85.29 &86.02& 84.4& 84.89 \\
64& 36.75& 58.31&   56.69&  78.81&  66.00&  63.82&  82.21&  80.19&  79.78&  \bf{85.25}& 83.95&  84.00 &83.39  &81.44  &81.08 \\
32& 15.07& 20.02 &  34.08&  78.16&  59.89&  60.05&  75.45&  72.89&  72.49&  \bf{84.28}& 79.01&  79.74 &76.58& 71.76&  70.42 \\
16& 6.36&13.94&   20.22&  68.56&  43.60&  43.44&  54.42&  50.04&  50.00&  \bf{81.40}& 64.99&  64.83& 67.22& 56.77&  55.71\\
8&  4.70& 3.85& 11.02&  45.99&  20.50&  20.5& 29.58&  25.53&  25.57 &\bf{52.39}&  32.62&  32.62 &40.96& 32.62&  32.58\\\Xhline{5\arrayrulewidth}
\end{tabular}
\label{table:classification}
\end{table*}

\begin{proposition} 
\label{proposition}
MOPS-Net is permutation-invariant.
 \end{proposition}

\noindent\textbf{Proof.}
Let $\mathbf{E}\in\mathbb{R}^{n\times n}$ be an arbitrary permutation matrix ($\mathbf{E}^\textsf{T}\mathbf{E}=\mathbf{E}\mathbf{E}^\textsf{T}=\mathbf{I}_n$), and  $\mathbf{\widetilde{P}}=\mathbf{E}\mathbf{P}$ be the permutated input. Denote by  $\mathbf{\widetilde{F}}\in\mathbb{R}^{n\times d}$ and $\mathbf{\widetilde{Q}}\in\mathbb{R}^{m\times 3}$ the corresponding high-dimensional point-wise features and sampled points when feeding $\mathbf{\widetilde{P}}$ into MOPS-Net, respectively. 

Notice that the non-linear function $\mathcal{H}(\cdot)$ extracts point-wise features with shared MLPs. Moreover, the global feature is aggregated via max-pooling, which is a symmetric function\footnote{See~\cite{qi2017pointnet} for more details about the permutation-invariant properties of max-pooling.}, hence
\begin{equation}
\mathbf{\widetilde{F}}=\mathcal{H}(\mathbf{\widetilde{P}})=\mathcal{H}(\mathbf{EP})=\mathbf{E}\mathcal{H}(\mathbf{P})=\mathbf{E}\mathbf{F}.
\end{equation}

Similarly, we obtain  
$\mathbf{\widetilde{S}}=\mathbf{E}\mathbf{S}$ due to the permutation-invariant property of softmax, implying
\begin{equation}
\mathbf{\widetilde{Q}}=\mathbf{\widetilde{S}}^\textsf{T}\mathbf{\widetilde{P}}=\mathbf{S}^\textsf{T}\mathbf{E}^\textsf{T}\mathbf{E}\mathbf{P}=\mathbf{S}^\textsf{T}\mathbf{P}=\mathbf{Q},
\end{equation}
indicating that  the sampled point set from $\mathbf{\widetilde{P}}$ is identical to those from $\mathbf{P}$, which completes the proof.


\subsection{Joint Training Loss}
\label{subsec:loss}
As analyzed in the objective function (\ref{eqn:relax}), two types of losses are needed to train MOPS-Net, i.e., the task loss $L_{task}(\cdot)$ and the subset loss $L_{subset}(\cdot, \cdot)$.  
Specifically, $L_{task}(\cdot)$ aims to promote the network to learn downsampled point clouds that are able to maintain the high performance for a specific task. Let $\mathscr{F}_T(\cdot)$ be the network for a typical task, which was trained with the original dense point cloud data, and we have
\begin{equation}
    L_{task}(\mathcal{Q})=L_T(\mathscr{F}_T(\mathcal{Q}), y^*),
\end{equation}
where $y^*$ is the corresponding ground-truth data for $\mathcal{Q}$. Specifically, $y^*$ will be the class label and the input point cloud when the task is classification and reconstruction, respectively. 

The subset loss $L_{subset}(\cdot, \cdot)$ regularizes the network to learn downsampled point clouds that are close to subsets of inputs, which is expressed as 
\begin{equation}
    L_{subset}(\mathcal{P},\mathcal{Q})=\frac{1}{m}\sum_{i=1,\dots,m}\min_{\mathbf{p}\in\mathcal{P}}||\mathbf{q}_i-\mathbf{p}||_2^2
\end{equation}

Therefore, the total loss $L_{total}(\cdot, \cdot)$ for end-to-end training of MOPS-Net is written as
\begin{equation}
    L_{total}(\mathcal{P},\mathcal{Q})=L_{task}(\mathcal{Q})+\alpha L_{subset}(\mathcal{P},\mathcal{Q}),
\end{equation}
where $\alpha>0$ balances the two terms. Figure \ref{fig:ablation} shows the effect of the value of $\alpha$ on performance.

\subsection{Flexible MOPS-Net for Arbitrary Ratios}
\label{subsec:FMOPS-Net}

In the previous sections, we construct MOPS-Net with a predefined sample size $m$, and a different network has to be trained for each $m$, which is tedious and unpractical for real-world applications. To solve this issue, we extend MOPS-Net and propose flexible MOPS-Net (FMOPS-Net), which is a single network that can achieve 3D point cloud downsampling with arbitrary sampling ratios after only one-time training.

Specifically, 
we consider learning a relatively large matrix $\mathbf{\widehat{S}}\in\mathbb{R}^{n\times m_{max}}$ with the same network architecture as MOPS-Net. Given an arbitrary sample size $m\leq m_{max}$, we select the $m$ left-most  columns of $\mathbf{\widehat{S}}$ to form the sampling matrix $\mathbf{S}_m\in\mathbb{R}^{n\times m}$, producing a point cloud $\mathcal{Q}_m$ with $m$ points according to Eq. (\ref{equ:regressed_point}).
Such a manner is equivalent to indirectly sorting the points of $\mathcal{P}$ according to their importance in a downsampled point cloud. 
To enable flexibility, we train  MOPS-Net by randomly picking the downsampled number $m\leq m_{max}$ and minimizing $ L_{total}(\mathcal{P}, \mathcal{Q}_{m})$ at each iteration.  Note that the computational complexity and the memory consumption during the inference phase are identical to Section~\ref{subsec:method_sampledP} by replacing $m$ with $m_{max}$.


\begin{figure}[t]
     \centering
         \includegraphics[width=0.4\textwidth]{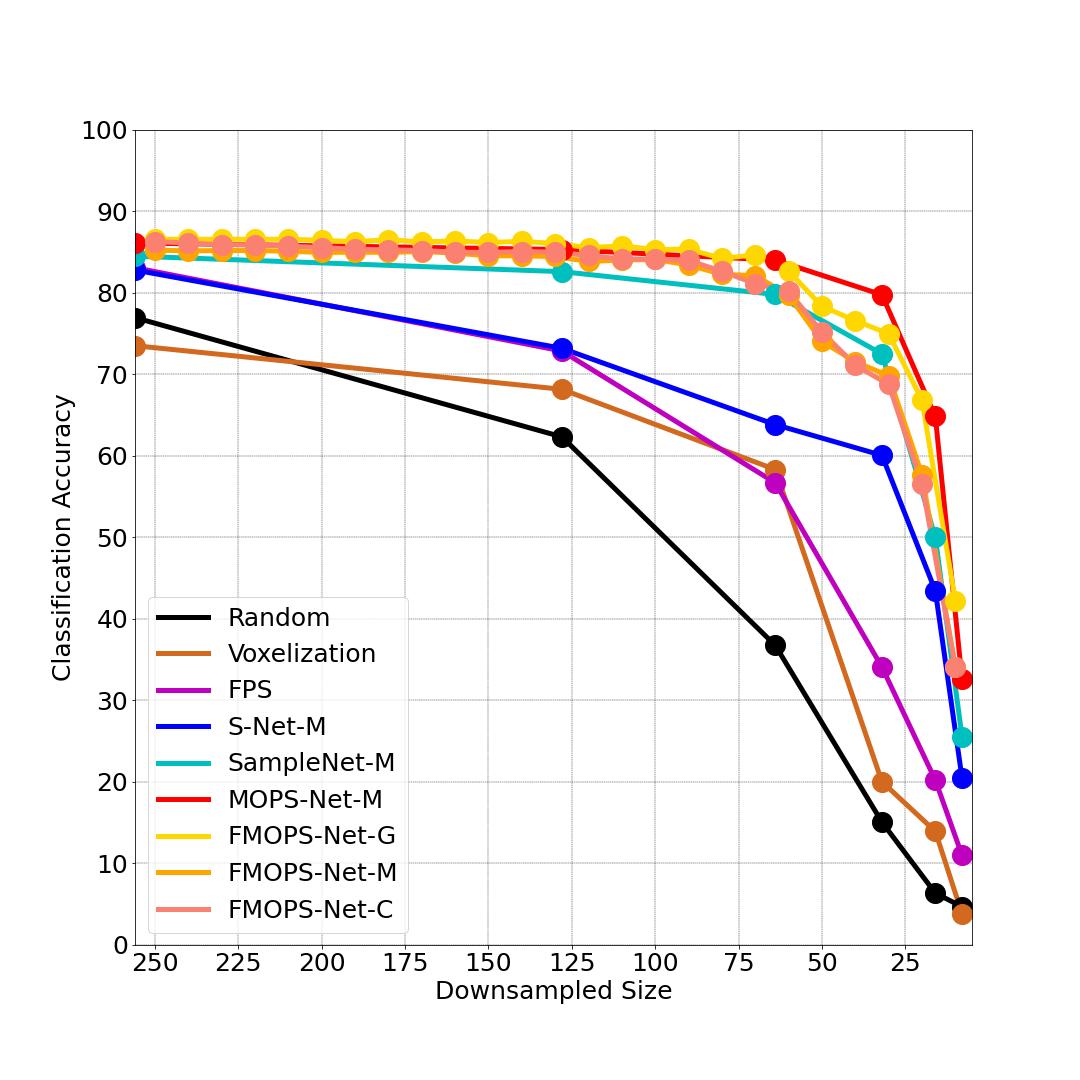}\vspace{-0.5cm}
        \caption{Quantitative comparisons of classification by different downsampling methods. Note that FMOPS-Net is applicable for arbitrary downsampled size.}
        \label{fig:cls_arbitrary}
\end{figure}

\begin{figure*}[t]
     \centering
 \includegraphics[width=7.5in]{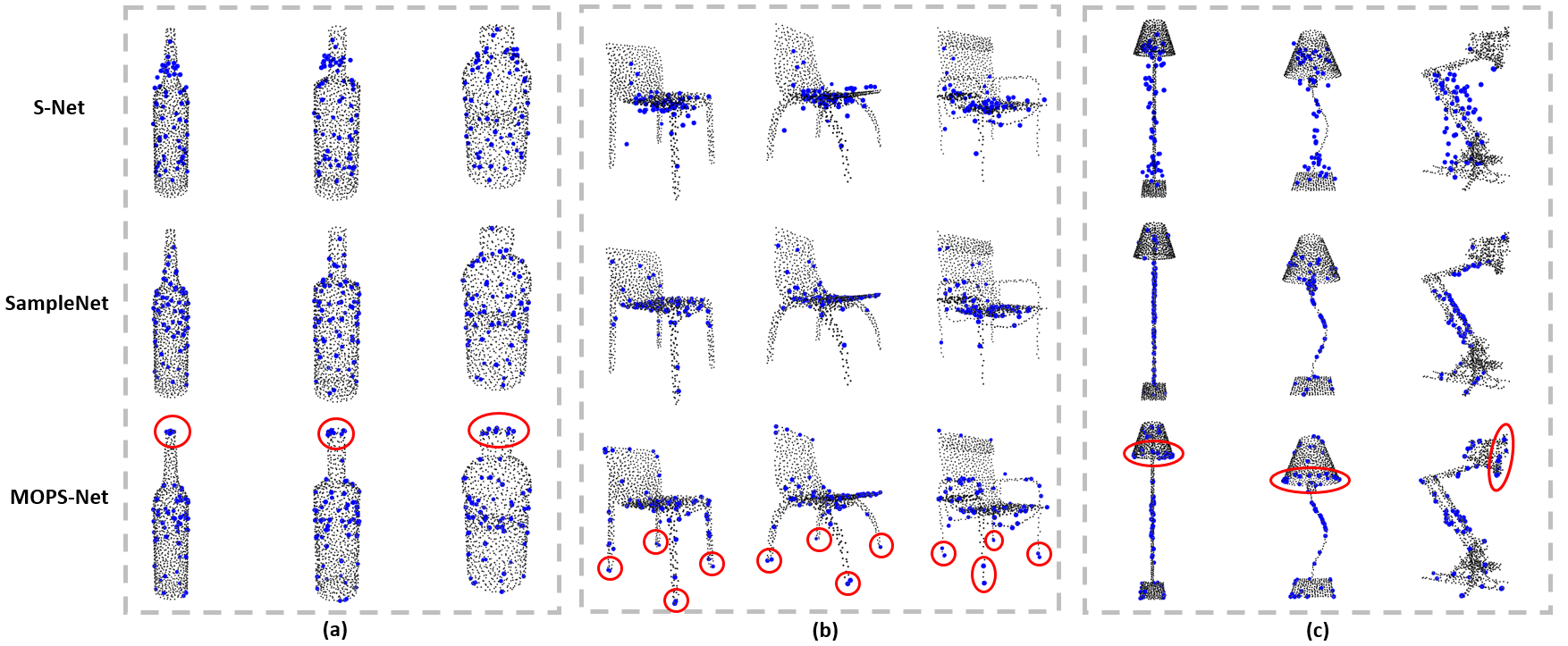}
        \caption{Visual comparisons of the generated point sets by the three deep learning-based task-oriented downsampling methods with $m=64$ over three classes: (a) Bottles  (b) Chairs  (c) Lamps. Prominent regions  within each class are highlighted in red. }
        \label{fig:cls_compare}
\end{figure*}

\begin{figure*}[htp!]
     \centering
         \includegraphics[width=0.9\textwidth]{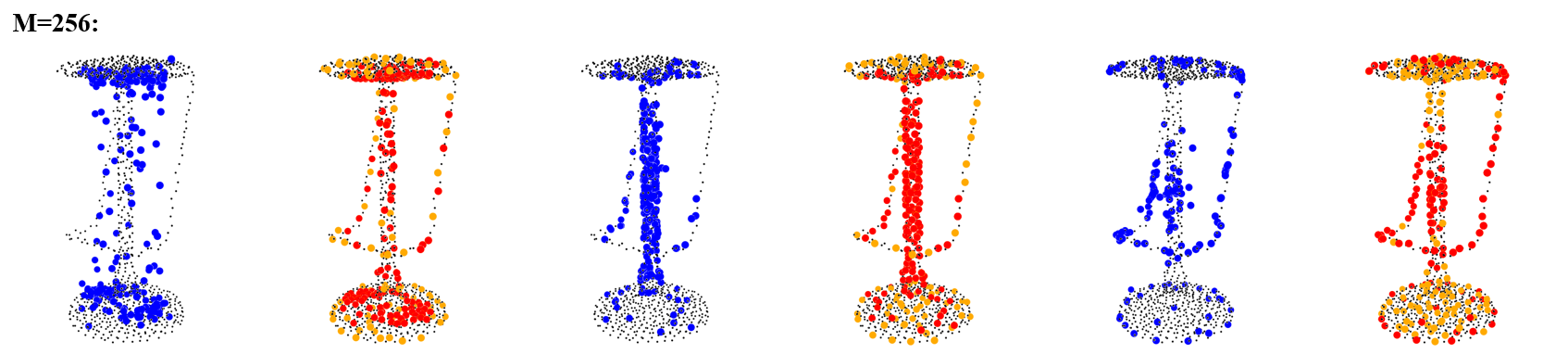}
         \includegraphics[width=0.9\textwidth]{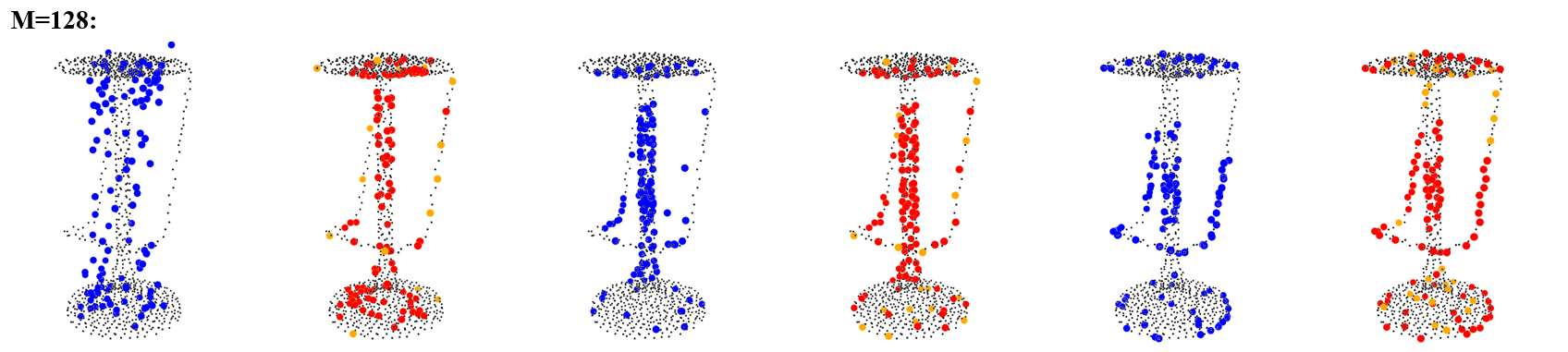}
         \includegraphics[width=0.9\textwidth]{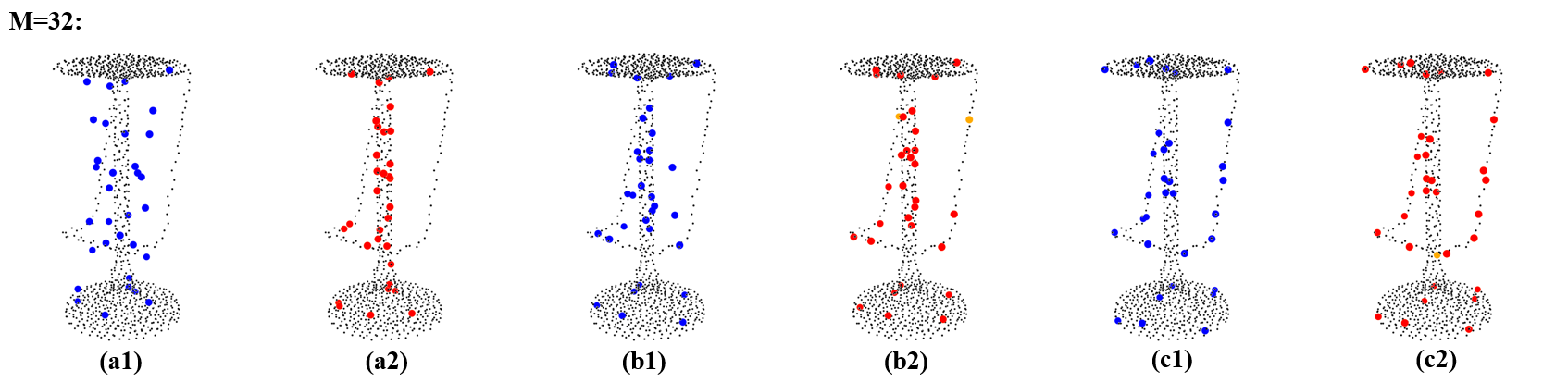}
        \caption{Visualization of sampled point clouds by different methods for $m=256$, $128$, and $32$. 
        (a1) \textcolor{blue}{generated} sets by S-Net; (a2) \textcolor{orange}{matched} and \textcolor{red}{completed} sets by S-Net; (b1) \textcolor{blue}{generated} sets by SampleNet; (b2) \textcolor{orange}{matched} and \textcolor{red}{completed} sets by SampleNet; (c1) \textcolor{blue}{generated} sets by MOPS-Net; (c2) \textcolor{orange}{matched} and \textcolor{red}{completed} sets by MOPS-Net. }
        \label{fig:cls_m}
\end{figure*}


\section{Experiments}
\label{sec:exp}
We validated the effectiveness of MOPS-Net and FMOPS-Net on  three typical tasks, i.e., point cloud classification, reconstruction, and registration. The task networks and experiment settings will be discussed in each section in detail. 
We used three widely used traditional downsampling methods, i.e., random sampling (RS), voxelization (Voxel) and farthest point sampling (FPS), as baselines. We also compared with S-Net~\cite{dovrat2019learning} and SampleNet~\cite{lang2020samplenet}, which are state-of-the-art deep learning-based task-oriented downsampling methods. Note that for S-Net, SampleNet, and MOPS-Net, a network was trained for a downsampling size.  For deep learning-based methods, we examined the performance of three types of downsampled sets, i.e., (1) \textbf{Generated (G)} sets: the point sets are generated directly by deep learning-based methods; (2) \textbf{Matched (M)} sets: the directly generated sets are post-processed via the matching operation, making the point sets be subsets of input ones; and  (3)  \textbf{Completed (C)} sets: the matched sets are further completed via FPS if their numbers of points are less than the specified value. In the following visual results, we visualized the \textcolor{blue}{\textbf{Generated (G)}}, \textcolor{red}{\textbf{Matched (M)}}, and \textcolor{orange}{\textbf{Completed (C)}} sets with \textcolor{blue}{blue}, \textcolor{red}{red}, and \textcolor{orange}{orange} colors, respectively, to distinguish them. 





\subsection{Classification-oriented Downsampling}
\label{subsec:class_exp}

\subsubsection{Classification of small-scale point clouds}
\label{subsec:class_exp_1k}

Following S-Net~\cite{dovrat2019learning} and SampleNet~\cite{lang2020samplenet}, we used the pre-trained PointNet vanilla~\cite{qi2017pointnet} performing on ModelNet40~\cite{wu20153d} as the classification task network. 
Note that the pre-trained PointNet vanilla trained on  point clouds with 1024 points achieves 87.1\% overall accuracy when classifying point clouds with 1024 points each to 40 categories.    
The task loss refers to the cross entropy between the predicted and ground-truth labels. During training of our MOPS-Net, we set $\tau_{min}=0.1$ and $\alpha=30$, and we initialized the learning rate to $5e^{-4}$ and  exponentially decreased it to $1e^{-5}$ within 250 epochs.


\textbf{Quantitative comparisons}. From Table \ref{table:classification}, we can observe that task-oriented downsampling methods, including S-Net, SampleNet and MOPS-Net, achieve much better performance than task-independent methods, including random sampling (RS), voxelization, and FPS. Note that the accuracy of S-Net drops significantly from generated sets to the matched subsets because the generative-based S-Net fails to obey the subset constraint. By replacing the repeated points in matched subsets by FPS points, the accuracy of completed subsets by  S-Net can be improved, especially for relatively large downsampled sizes $m=$256 and 512. 

SampleNet improves S-Net by projecting the generated sets to nearest neighbors in original point clouds, and thus can minimize the performance gap between the generated set and matched subset. However, because of the additional restriction for projection, the accuracy of generated points by SampleNet is inferior to that by S-Net for relatively small downsampled sizes $m=8,16,32$. Moreover, SampleNet fails to generate meaningful points for $m=512$, and it mainly relies on additional FPS postprocessing to obtain the completed set which can result in comparable performance.

The proposed MOPS-Net consistently achieves the best performance over all cases, which is credited to real sampling process-like modeling of MOPS-Net. 
The flexibility of FMOPS-Net is achieved at the cost of slight degradation of the accuracy of  MOPS-Net; 
however, FMOPS-Net still outperforms S-Net and SampleNet under almost all cases. 
Figure~\ref{fig:cls_arbitrary} further demonstrates the flexibility and advantage of FMOPS-Net by showing the accuracy of more downsampling sizes. 

\textbf{Visual comparisons}. 
 Figure~\ref{fig:cls_compare} visually illustrates sampled point clouds  ($m=64$) by the three deep learning-based task-oriented downsampling methods, i.e. S-Net, SampleNet, and MOPS-Net, on three classes. 
From Figure~\ref{fig:cls_compare}, it can be seen that 
the two generative-based S-Net and SampleNet tend to generate points close to the centers of shapes and fail to capture prominent regions. By contrast, our MOPS-Net can successfully select points near the contours of shapes and focus on prominant regions. Besides, for different point clouds of a typical class, MOPS-Net focus on selecting points corresponding to identical semantics, e.g. the bottleneck of bottles, the legs of chairs, and the lampshade of lamps. These observations also explain why the downsampled point clouds by MOPS-Net can be classified with higher accuracy than S-Net and SampleNet. 

Besides, Figure~\ref{fig:cls_m} visualizes the \textcolor{blue}{generated}, \textcolor{red}{matched}, and \textcolor{orange}{completed} sets by S-Net, SampleNet, and MOPS-Net under various $m$.
where it can be seen that S-Net fails to directly generate downsampled points close to the input points once the shape is complex, and SampleNet tends to directly generate points clustered around the shape center and omit the footrest of the stool, which is assumed to be important for shape identification. By contrast, our MOPS-Net can consistently capture these discriminative regions for any downsampled size.


\subsubsection{Classification of large-scale point clouds}

To demonstrate the potential of MOPS-Net for processing large-scale point clouds, we further applied MOPS-Net for downsampling point clouds with 10,000 and 100,000 points each. We first pre-trained a PointNet-vanilla classifier on ModelNet40 with 100,000 points each model. The other settings, including the optimized loss function, training epoch and the training strategies, were identical to the experiments in Section~\ref{subsec:class_exp_1k}. The pre-trained classifier can achieve 90.11\% overall accuracy.
By fixing the pre-trained classifier, we then trained MOPS-Net for downsampling point clouds with 10,000 points each. The experiment settings were kept identical to our previous experiments on small-scale point clouds. As MOPS-Net is built upon point-wise MLPs and the Softmax operator, which are independent of the number of input points, we can directly apply MOPS-Net trained on 10,000 points for downsampling larger-scale point clouds with 100,000 points each.

Table~\ref{table:cls_10w} lists the classification accuracy of  MOPS-Net for downsampling 10,000 points and 100,000 points. The high classification accuracy demonstrates the ability for the proposed MOPS-Net on downsampling large-scale point clouds. In particular, MOPS-Net trained on point clouds with 10,000 points each can be successfully extended to downsample larger-scale point clouds with 100,000 points each, without any modification or fine-tune, which also demonstrates its flexibility.

\begin{table}[t]
\centering
\caption{Comparisons of the classification accuracy on downsampling large-scale point clouds. The larger, the better.}\vspace{-0.1in}
\begin{tabular}{c||ccc|ccc}\Xhline{5\arrayrulewidth}
 & \multicolumn{3}{c|}{$n=10,000$} & \multicolumn{3}{c}{$n=100,000$} \\  
  $m$&  \makebox[3em]{RS}&  \makebox[3em]{FPS} & \makebox[3.5em]{MOPS-Net} & \makebox[3em]{RS}&  \makebox[3em]{FPS} & \makebox[3.5em]{MOPS-Net}\\\Xhline{2\arrayrulewidth}
1024&   84.97 &88.65& \bf{90.24}& 84.93&  88.98&  \bf{89.79} \\
512&  77.23 &85.69& \bf{90.00}& 76.86&  85.86 &\bf{89.18}\\
256 & 65.84 &79.67& \bf{87.40}  &64.83  &79.78& \bf{89.10}\\
\Xhline{5\arrayrulewidth}
\end{tabular}
\label{table:cls_10w}
\end{table}

\subsection{Reconstruction-oriented Downsampling}
\label{subsec:rec_exp}

\subsubsection{Reconstruction of small-scale point clouds}
\label{subsubsec:exp_recsmall}
In this scenario, we followed the settings of S-Net and SampleNet to evaluate our method. 
The task network $\mathscr{F}_T(\cdot)$ was achieved by a  pre-trained MLP-based reconstruction network\cite{achlioptas2018learning}, where a 3-layer MLP of size $[256,~256,~1024\times3]$ is utilized to reconstruct point clouds with 1024 points from a 128-dimensional global feature. 
We obtained the 128-dimensional global features of downsampled point clouds by applying max-pooling on the pointwise features extracted by a 5-layer MLP of size $[64,~128,~128,~256,~128]$. 
A single class of ShapeNetCore~\cite{chang2015shapenet} was used for training and testing. 
The task loss was set as the combination of the Chamfer distance (CD) and earth-mover distance (EMD).  
During training of our MOPS-Net, we 
set $\tau_{min}=0.5, \alpha=0.2$ and initialized the learning rate to  $5e^{-4}$ and exponentially decreased it to $1e^{-5}$ within 250 epochs.
We quantitatively measured the reconstruction performance of different downsampling methods using the normalized reconstruction error (NRE) for CD and EMD, which are defined as 
\begin{equation}
    \textsf{NRE}_\textsf{CD}(\mathcal{Q}, \mathcal{P})=\frac{\textsf{CD}(\mathcal{P}, \mathscr{F}_T(\mathcal{Q}))}{\textsf{CD}(\mathcal{P}, \mathscr{F}_T(\mathcal{P}))}.
    \label{eqn:nre}
\end{equation}
\begin{equation}
    \textsf{NRE}_\textsf{EMD}(\mathcal{Q}, \mathcal{P})=\frac{\textsf{EMD}(\mathcal{P}, \mathscr{F}_T(\mathcal{Q}))}{\textsf{EMD}(\mathcal{P}, \mathscr{F}_T(\mathcal{P}))},
    \label{eqn:nre}
\end{equation}
where $\textsf{CD}(\cdot)$ and $\textsf{EMD}(\cdot)$ compute  the CD and EMD, respectively.  The values of $\textsf{NRE}_{\textsf{CD}}$ and $\textsf{NRE}_{\textsf{EMD}}$ are lower bounded by 1, and the smaller, the better. 


\begin{figure}[t]
     \centering  
    \subfigure[]{ \includegraphics[width=1.7in]{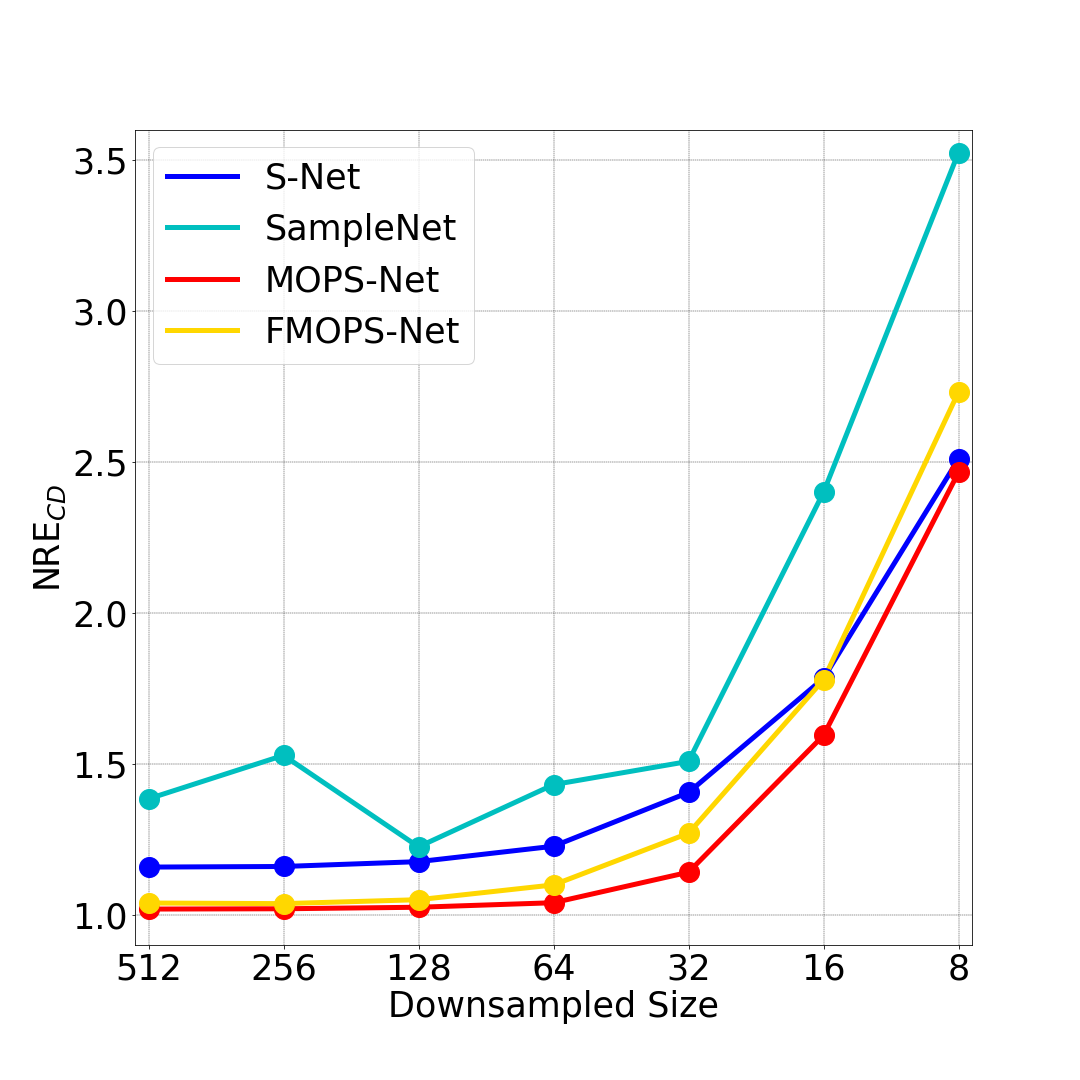}                  \includegraphics[width=1.70in]{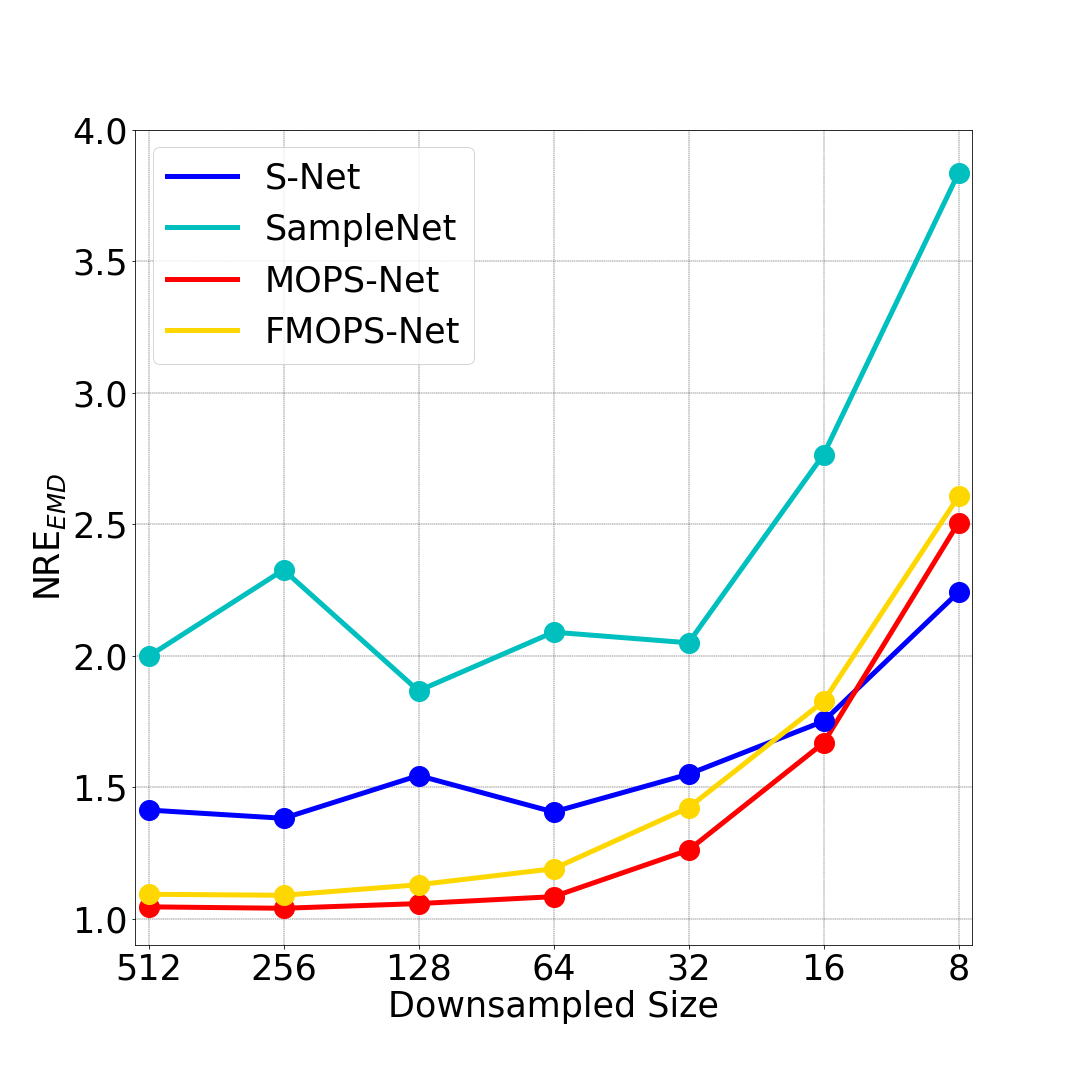}} \vspace{-0.6cm} 
    \subfigure[]{ \includegraphics[width=1.7in]{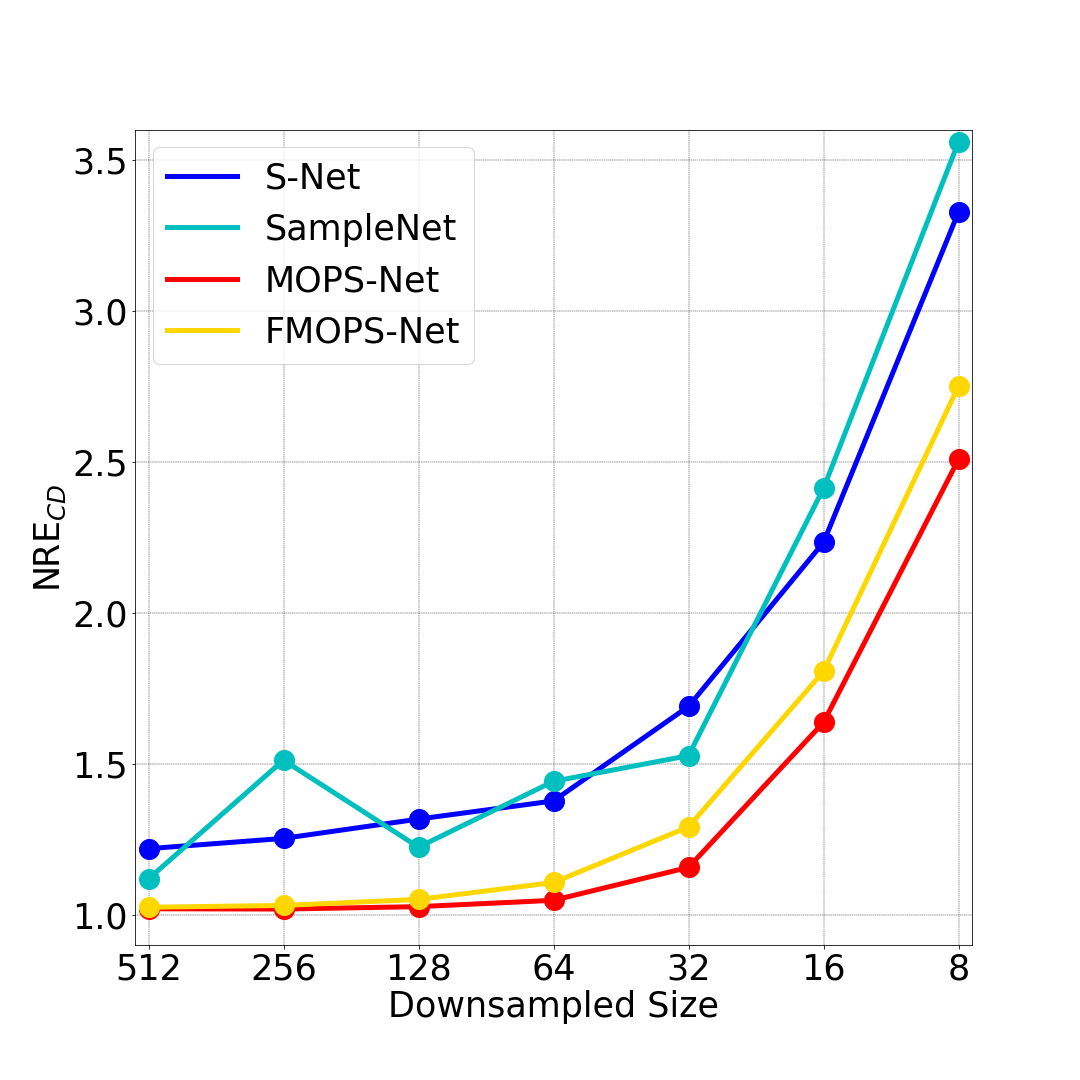}  \vspace{-0.6cm}  
               \includegraphics[width=1.7in]{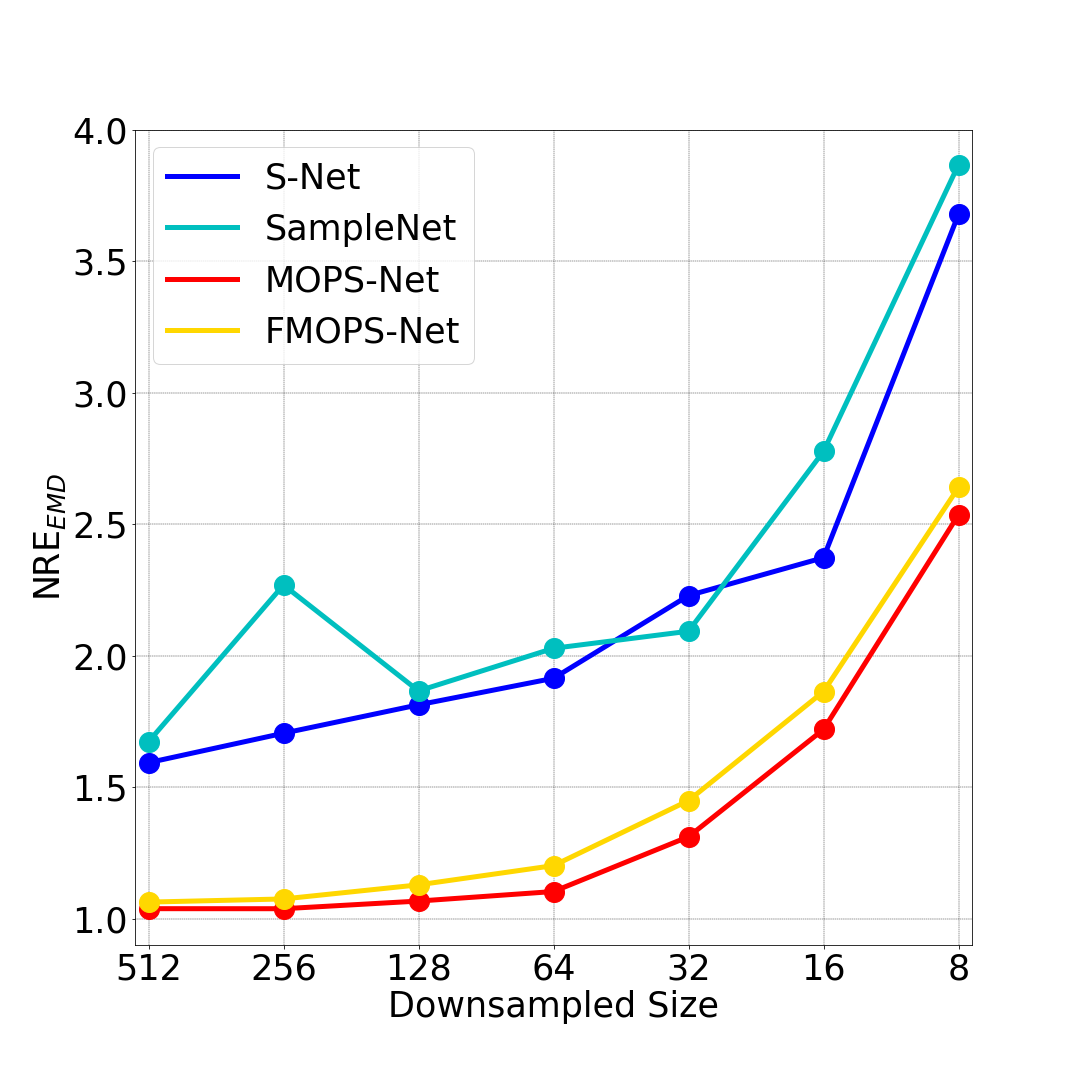} } 
     \subfigure[]{\includegraphics[width=1.7in]{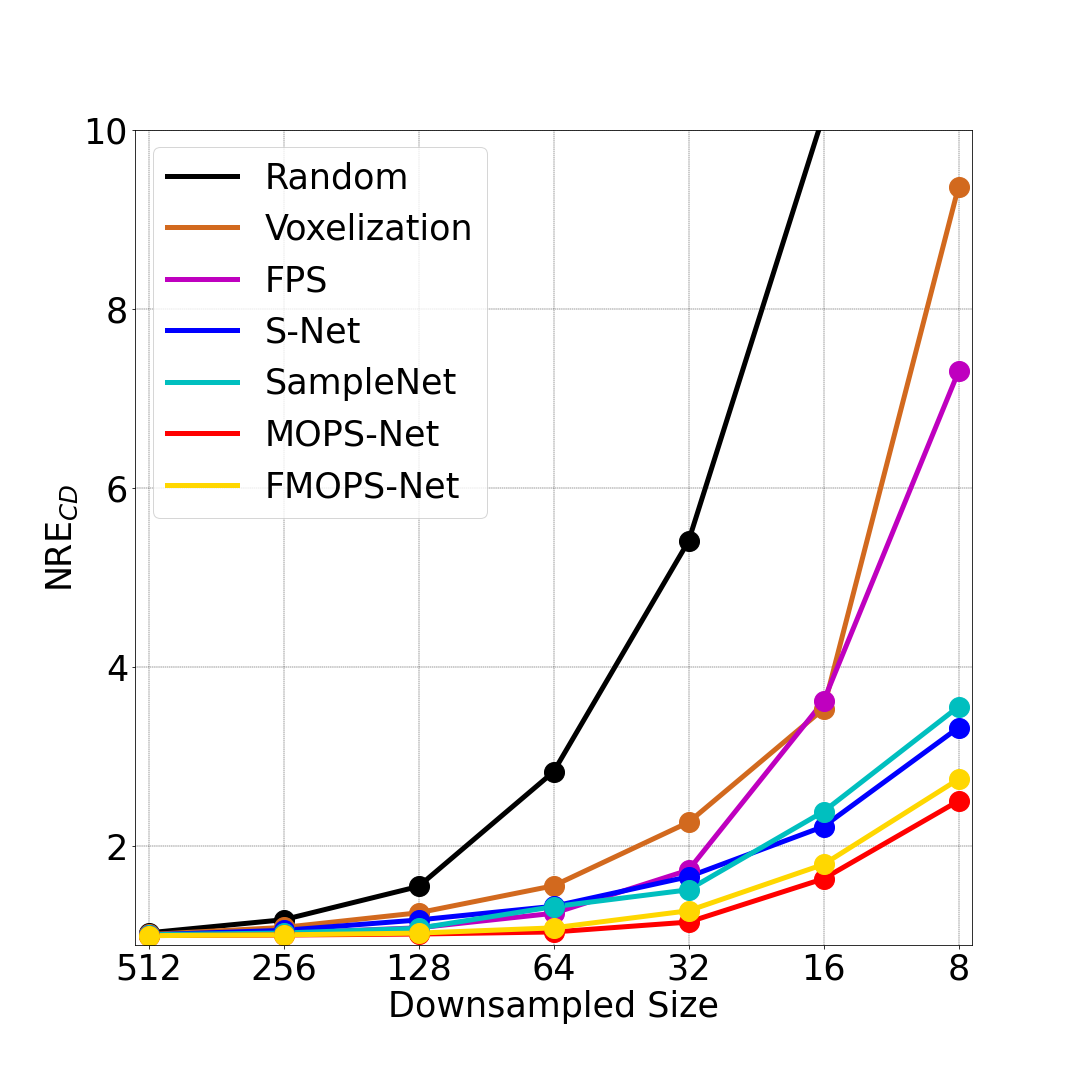}
      \includegraphics[width=1.7in]{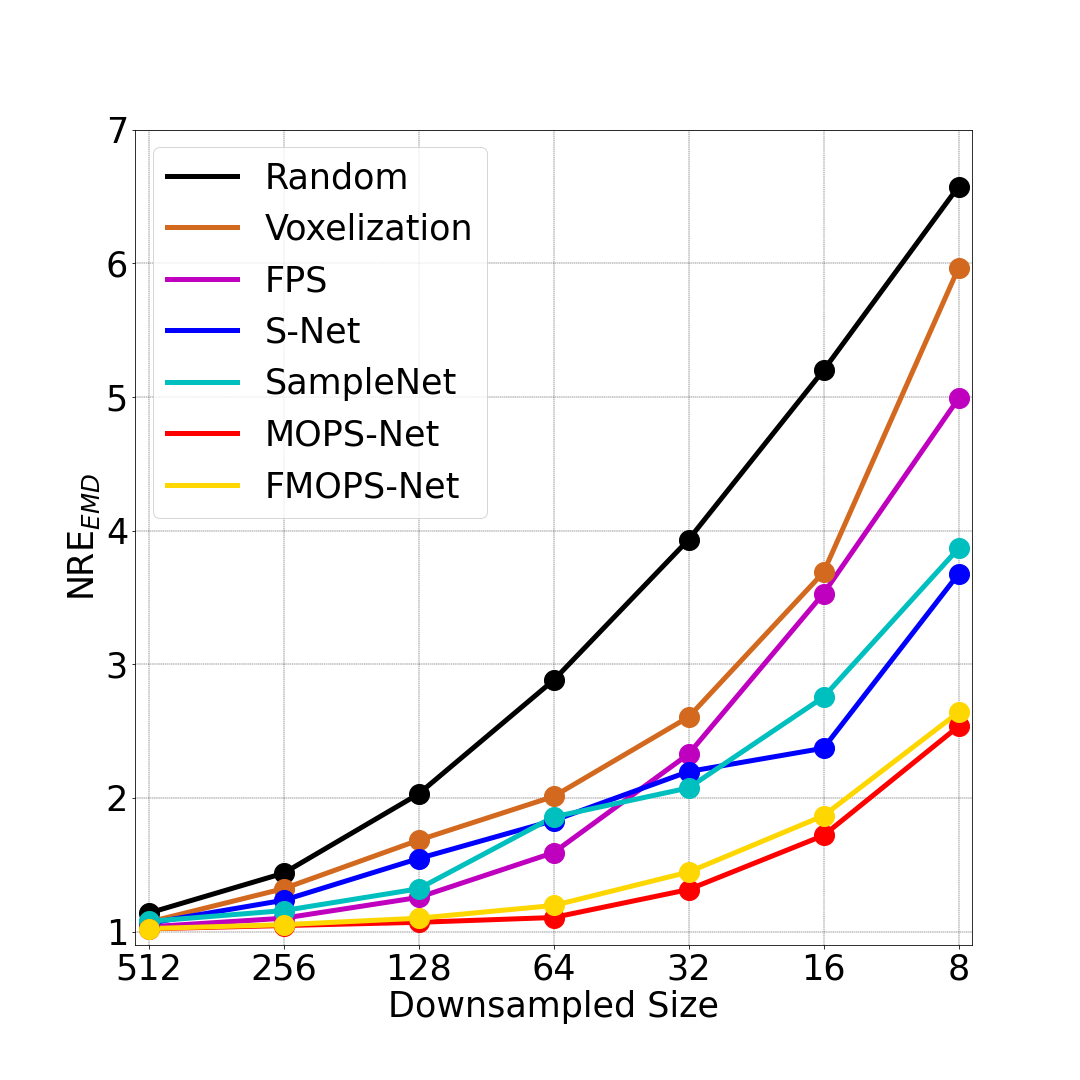}}  \vspace{-0.4cm}
    \caption{Quantitative comparisons of the distortion of reconstructed dense point clouds from the corresponding downsampled ones by different downsampling methods. Reconstruction from (a) generated sets (b) matched sets (c) completed sets. 
    }
    \label{fig:nre}
\end{figure}

\begin{figure*}[htp!]
     \centering
         \includegraphics[width=\textwidth]{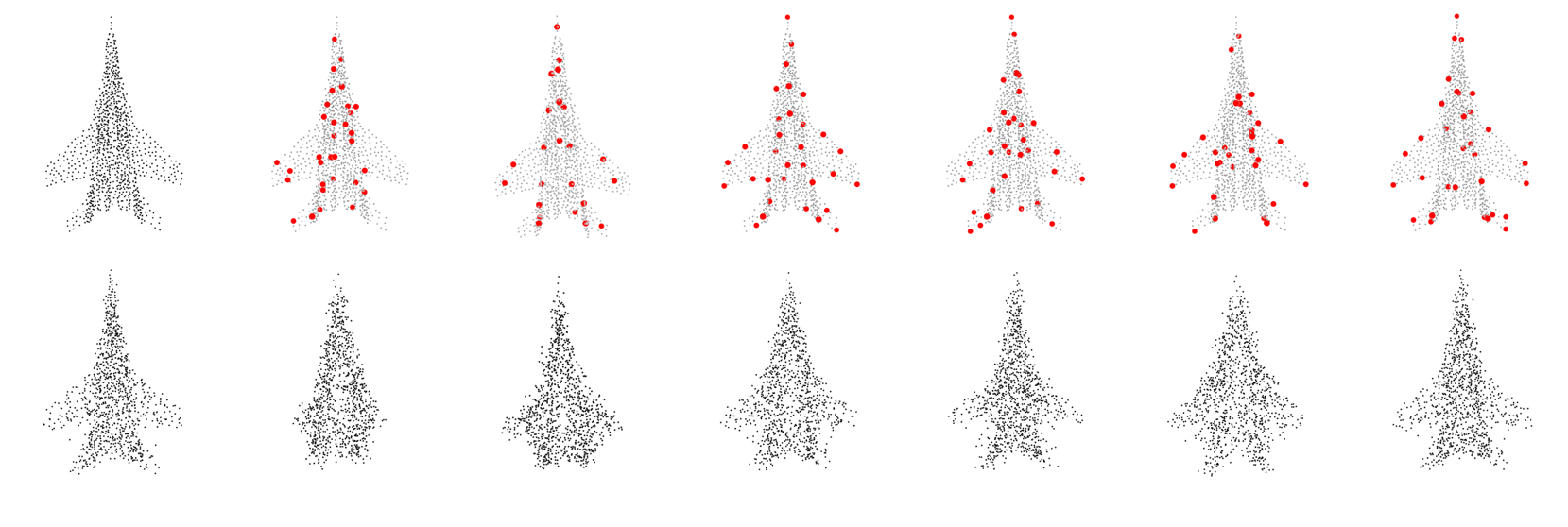}
         \includegraphics[width=\textwidth]{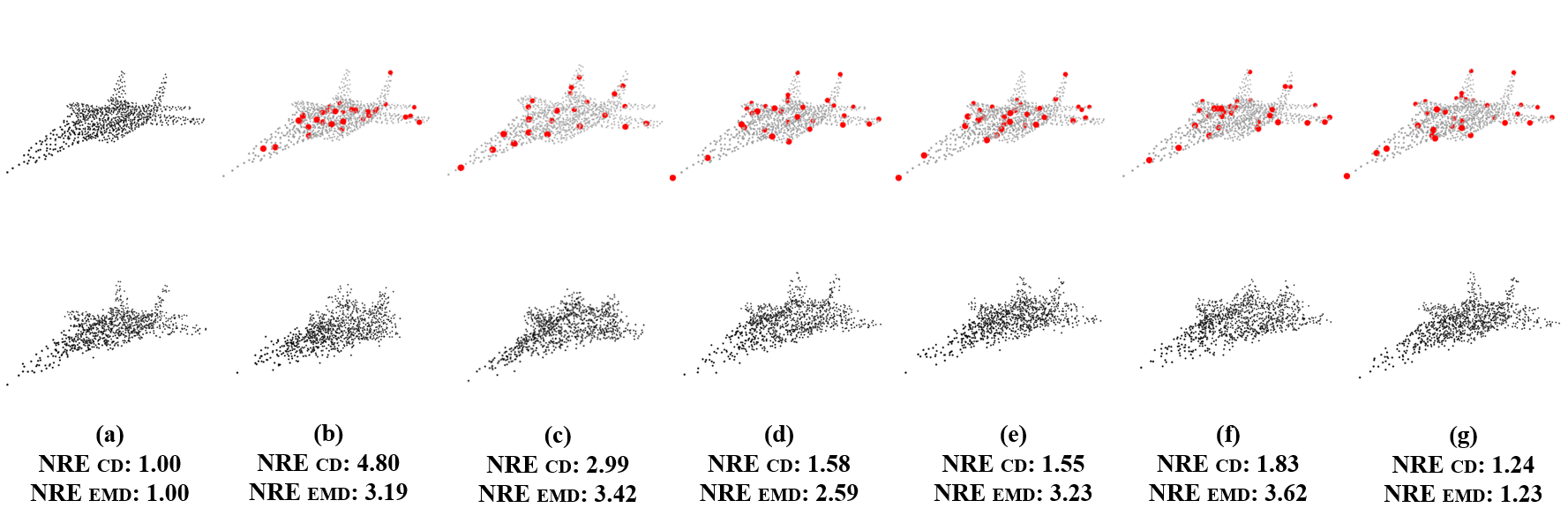}
        \caption{Visual comparisons of downsampled point clouds (1$^{st}$ and 3$^{rd}$ rows) and reconstructed point clouds (2$^{nd}$ and 4$^{th}$ rows) by different downsampling methods with $m=32$. (a) Original point clouds and corresponding reconstructions. (b)
        Random sampling; (c) Voxelization; (d) FPS; (e) S-Net; (f) SampleNet; (g) MOPS-Net.}
        \label{fig:reconst_compare}
\end{figure*}

\textbf{Quantitative comparisons}.
Figures~\ref{fig:nre}(a) and (b) show the $\textsf{NRE}_{\textsf{CD}}$ and $\textsf{NRE}_{\textsf{EMD}}$ values of the reconstructed point clouds from generated and matched sets by different donwsampling methods under various downsampling sizes, where it can be seen that  
except the extremely small downsampled size ($m=8$ and $16$), 
our MOPS-Net and FMOPS-Net produce much lower distortion than S-Net and SampleNet.  For the reconstruction from completed sets shown in Figure~\ref{fig:nre}(c), S-Net and SampleNet are even worse than FPS when $m>64$. The reason may be that FPS is able to produce uniformly distributed downsampled points; however, the uniform distribution cannot be guaranteed by S-Net and SampleNet as they are generative methods. 
However, our MOPS-Net and FMOP-Net consistently achieve the best and second best performance under all downsampled sizes, respectively, and FMOP-Net is  even comparable to MOPS-Net. 



\begin{figure*}[htp!]
     \centering
         \includegraphics[width=0.9\textwidth]{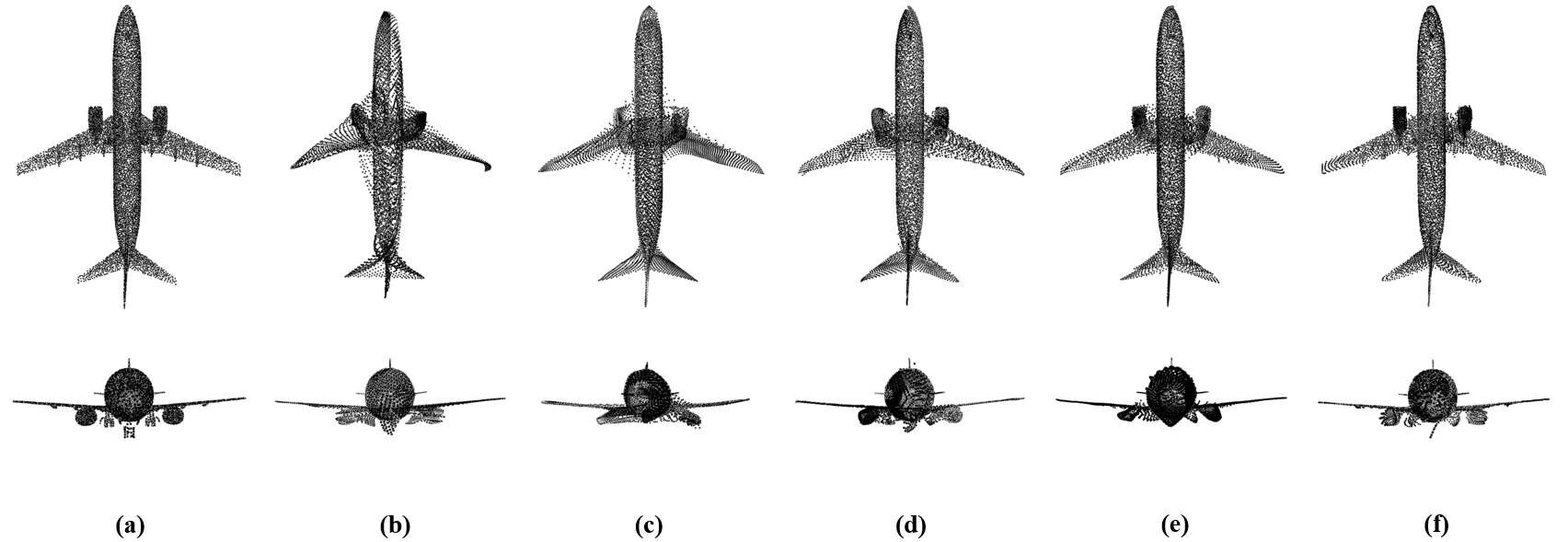}
        \caption{Visual comparisons of the reconstructed large-scale point clouds via different reconstruction frameworks. (a) Original point cloud with 10, 000 points; Reconstructed by (b)  FoldingNet; (c) M-FoldingNet ($M=4$); (d)  M-FoldingNet ($M=8$); (e) M-FoldingNet ($M=16$); (f) M-FoldingNet ($M=64$).}
        \label{fig:reconst_10k}
\end{figure*}

\textbf{Visual comparisons}.   Figure~\ref{fig:reconst_compare} visually compares the reconstructed dense point clouds from sparse ones obtained by different downsampling methods, 
where it can be seen that the reconstructed point clouds from our MOPS-Net are much better than those from other downsampling methods and are much closer to ground-truth ones. This advantage is  
is credited to that the downsampled points by our MOPS-Net can well capture the contour and salient features of 3D shapes.

\begin{figure}[htp!]
     \centering
         \includegraphics[width=0.5\textwidth]{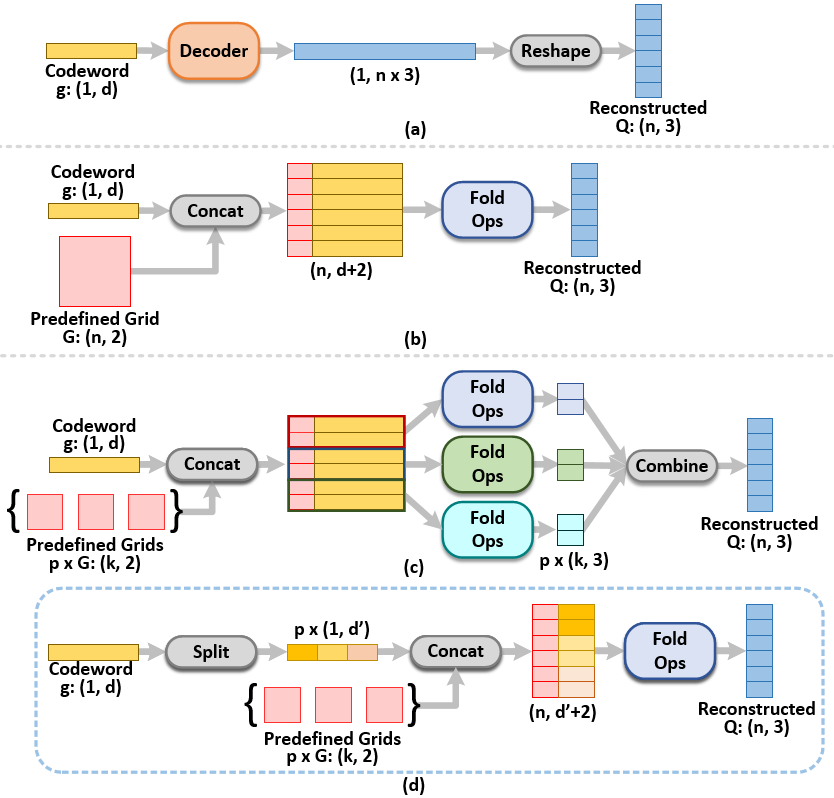}
        \caption{Illustration of the differences between different frameworks for point cloud reconstruction. (a) MLP-based; (b) FoldingNet  ~\cite{yang2018foldingnet}; (c) AtlasNet~\cite{groueix2018papier}; (d) Proposed M-FoldingNet. Note that the multiple folding operators of the AtlasNet are independent. 
        For FoldingNet, AtlasNet, and M-FoldingNet, the number of reconstructed points depends on the dimensions of the 2D grids.}
        \label{fig:mfold}
\end{figure}

\begin{table*}[htp!]
\centering
\caption{Quantitative comparisons of different reconstruction frameworks. 
The smaller, the better. Model sizes were measured under the same parameter precesion.}\vspace{-0.1in}
\begin{tabular}{c|c||cc|cccccc}\Xhline{5\arrayrulewidth}
&&&&\multicolumn{6}{c}{M-FoldingNet} \\
points & metric & \makebox[4em]{MLP-based} & \makebox[7em]{FoldingNet~\cite{yang2018foldingnet}} &  \makebox[4em]{$M=4$} & \makebox[4em]{$M=8$} & \makebox[4em]{$M=16$} & \makebox[4em]{$M=32$} & \makebox[4em]{$M=64$} & \makebox[4em]{$M=128$}  \\\Xhline{2\arrayrulewidth}
1024& CD&13.81& 18.48&  \bf{11.84}& 12.69&  13.46&  13.34 & 13.70 & 16.46 \\
& EMD&  \bf{154.26}&  359.78&   175.07&209.13&  210.69& 210.69 & 198.66 & 222.89\\
& Model Size & 4.1MB & 4.7 MB & 3.1MB & 2.9 MB & 2.8MB &2.8MB & 2.8MB & 2.8MB \\
\Xhline{2\arrayrulewidth}
10000&  CD& NA& 1.25& 0.84& 0.58& 0.49& 0.42& \bf{0.36}&  0.36 \\
& EMD&  NA  &18.8&  18.3& 11.27&  7.31& 4.86& \bf{2.56}&  2.70\\
& Model Size &NA & 12.0MB & 5.4 MB & 4.4MB & 3.9 MB & 3.6MB & 3.5MB &  3.5MB \\
\Xhline{5\arrayrulewidth}
\end{tabular}
\label{table:multi_foldnet}
\end{table*}

\subsubsection{Reconstruction of large-scale point clouds}
To demonstrate the ability of our MOPS-Net on large-scale point clouds,  we also examined MOPS-Net on 
 point clouds with 40,960 points each.
Unfortunately, the MLP-based reconstruction framework (see  Figure~\ref{fig:mfold}(a)) employed in Section \ref{subsubsec:exp_recsmall} cannot well adapt to large-scale point cloud reconstruction because the network size is 
linearly proportional to the number of output points. 
To this end, we also propose a new framework for reconstructing large-scale point clouds, whose network size is independent 
of the output point number. 

The proposed framework for large-scale point cloud reconstruction, namely Multi-FoldingNet (M-FoldingNet), is motivated by FoldingNet~\cite{yang2018foldingnet} shown in Figure~\ref{fig:mfold}(b). As illustrated in Figure~\ref{fig:mfold}(d), instead of concatenating the global feature to the 2D coordinates of a single regular grid, we first segment such a global feature into a set of $M$ local features with a smaller and equal feature dimension, and it is expected that each local feature encodes the high-level semantic information of a local patch on a point cloud . We then  concatenate each local feature to the coordinates of a 2D regular grid separately, which are fed into a shared folding operator.  Note that the number of output points can be varied by adjusting the dimensions of the 2D regular grids. 
Compared with FoldingNet, which can be thought of as a special case of M-FoldingNet with $M=1$, M-FoldingNet with fewer network parameters can allow the folding operator to focus on local regions which are easier to be reconstructed. 
Although  AtlasNet~\cite{groueix2018papier} illustrated in Figure~\ref{fig:mfold}(c) also realizes reconstruction in a local manner, it adopts multiple independent folding operators, leading to the significant increase of network parameters, compared with FoldingNet.

We first evaluated and compared the proposed M-FoldingNet
with other reconstruction frameworks on both small-scale point clouds (1024 points) and large-scale point clouds (10,000 points each)\footnote{Here we used point clouds with 10,000 points each to enable the application of the MLP-based framework.}. The settings, including the dataset, the generation of codewords/global features, 
and the task loss, were kept identical to the MLP-based reconstruction framework 
in Section \ref{subsubsec:exp_recsmall}. 
For fair comparisons, we used an identical codeword dimension for all reconstruction frameworks, i.e., the codeword dimension equals to $d=128$ (resp. $d=512$) for reconstructing point clouds with $n=1024$ (resp. $n=10,000$) points. As listed in Table~\ref{table:multi_foldnet}, we can observe that the proposed M-FoldingNet can achieve better performance than FoldingNet. Specifically, M-FoldingNet with $M=4$ can achieve best performance on the reconstruction of small-scale point clouds. The reconstruction quality decreases with $M$ increasing because the segmented local features 
have a limited dimension to fully embed local part information. 
As expected, more pieces of local features are needed to achieve best reconstruction performance. Besides, our M-FoldingNet is more compact than FoldingNet and MLP-based. 
Figure~\ref{fig:reconst_10k} visually compares the reconstructed point clouds by M-FlodingNet and FoldingNet, 
which also demonstrates the superiority of the proposed reconstruction framework 


\begin{table}[t]
\centering
\caption{Quantitative comparisons of the distortion of reconstructed point clouds from downsampled sparse point clouds by different methods. The original point clouds consist of 40,960 points each. The smaller, the better. }\vspace{-0.1in}
\begin{tabular}{c||ccc|ccc}\Xhline{5\arrayrulewidth}
 && $\textsf{NRE}_{\textsf{CD}}$ &&& $\textsf{NRE}_{\textsf{EMD}}$& \\
$m$ &  \makebox[3em]{RS}&  \makebox[3em]{FPS} &  \makebox[3em]{MOPS}&  \makebox[3em]{RS}&  \makebox[3em]{FPS} &  \makebox[3em]{MOPS} \\\Xhline{2\arrayrulewidth}
4096& 1.37  &1.33&  \bf{1.06}&  3.93& 3.23& \bf{1.57} \\
1024& 4.00  &2.53&  \bf{1.35}&  12.10&  8.97  &\bf{3.05}\\
256 & 26.77 &7.24&  \bf{2.71} &45.53  &19.13& \bf{7.40}\\
\Xhline{5\arrayrulewidth}
\end{tabular}
\label{table:reconstruction_10k}
\end{table}

\begin{figure*}[htp!]
     \centering
         \includegraphics[width=0.8\textwidth]{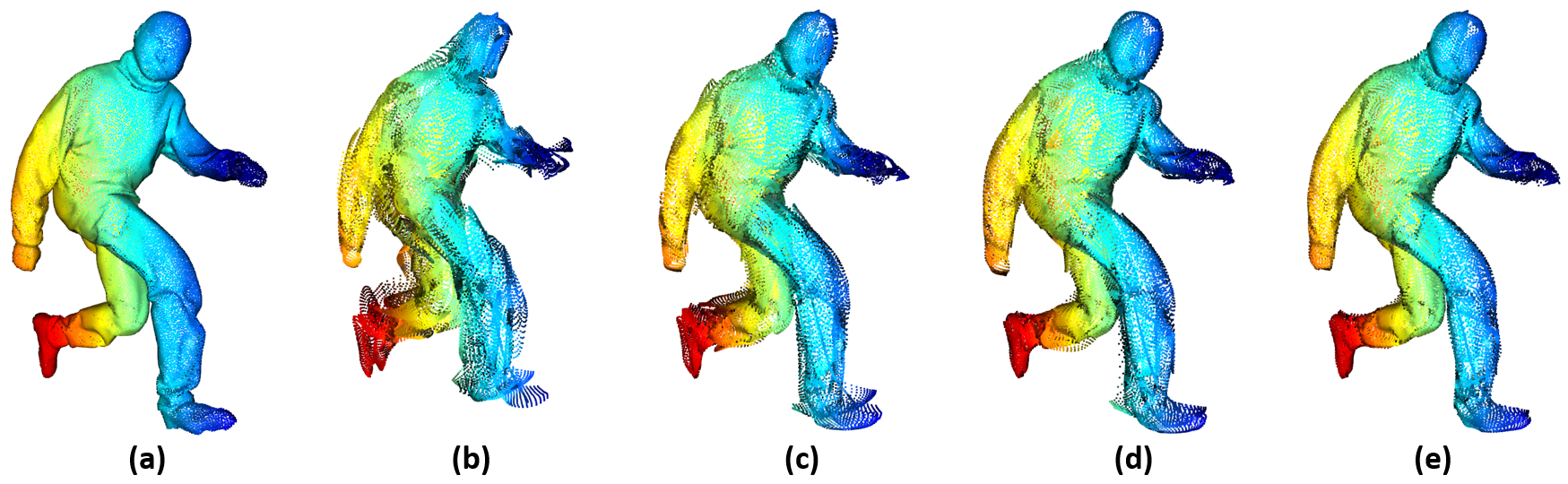}
        \caption{Visual comparisons of reconstructed large-scale point clouds by different downsampling methods with $m=256$. (a) Original 40,960 real scanned data. Reconstructed dense point clouds from 256 points sampled by (b) Random sampling; (c) FPS; (d) MOPS-Net; (e) Reconstructed dense point cloud by extracting the codeword from the original point cloud. Colors are assigned by the pointwise depths for better visualization.}
        \label{fig:scan}
\end{figure*}

We evaluated the performance of MOPS-Net for downsampling  real scanned data with 40,960 points each~\cite{vlasic2008articulated}, where the pre-trained M-FoldingNet ($M=128$ and $d=2048$ (or $d'=16$)) was used as the task network. 
The settings of MOPS-Net were the same as those in Section \ref{subsubsec:exp_recsmall}. 
Table~\ref{table:reconstruction_10k} quantitatively compares the distortion of reconstructed point clouds from the downsampled sparse point clouds by random sampling, FPS, and MOPS-Net 
\footnote{Note that we did not provide the results of S-Net and SampleNet in this experiment because it is difficult to tune the hyper-parameters contained in the loss functions of S-Net and SampleNet for obtaining satisfied results due to the large-scale point clouds. To ensure the correctness of our paper, we omitted their results. Besides, in Section \ref{subsec:ablation}, we quantitatively analyzed the effect of the hyper-parameters contained in the loss function of SampleNet on reconstruction performance.}. 
From Table~\ref{table:reconstruction_10k}, we can see that the $\textsf{NRE}_{\textsf{CD}}$ and $\textsf{NRE}_{\textsf{EMD}}$ values of the reconstruction from the downsampled points by MOPS-Net are much smaller than those of the reconstruction from downsampled points by RS and FPS under all cases, 
which demonstrates the ability of MOPS-Net in handling large-scale point clouds. Besides, in Figure~\ref{fig:scan}, we visualized the reconstructed dense point clouds from downsampled  $m=256$ points via different methods. The high quality of reconstructed 40,960 dense point clouds from MOPS-Net demonstrates the superiority of the proposed method for downsampling large-scale point clouds. The memory and computational complexity for downsampling large-scale point clouds can also be found in Section~\ref{subsec:complexity}. Last but not least, for a very large-scale point cloud in practice, a promising solution is to partition it to several regions with smaller points each and then downsample the regions in parallel or sequentially.

\begin{table*}[t]
\centering
\caption{Quantitative comparisons of MREs of donwsampled point clouds by different methods used for registration. The smaller, the better.}\vspace{-0.1in}
\begin{tabular}{c||c|c|c|ccc|ccc|ccc|ccc}\Xhline{5\arrayrulewidth}
   &\makebox[2.7em]{RS}&\makebox[2.7em]{Voxel}&\makebox[2.7em]{FPS} &&\makebox[2.7em]{S-Net ~\cite{dovrat2019learning}}&&&\makebox[2.7em]{SampleNet \cite{lang2020samplenet}}&&&\makebox[2.7em]{MOPS-Net}& &&\makebox[2.7em]{FMOPS-Net} &\\
      $m$ & & & &\makebox[2.7em]{G}&\makebox[2.7em]{M}&\makebox[2.7em]{C}&\makebox[2.7em]{G}&\makebox[2.7em]{M}&\makebox[2.7em]{C}&\makebox[2.7em]{G}&\makebox[2.7em]{M}&\makebox[2.7em]{C} &\makebox[2.7em]{G}&\makebox[2.7em]{M}&\makebox[2.7em]{C}     \\\Xhline{2\arrayrulewidth}
64& 17.47&  10.37 & 9.98& 10.93&  12.83&  13.26&8.30& 8.69& 8.29& 8.58& 8.56& 8.00& 7.94& 8.87& 8.84\\
32& 26.95&  14.46& 13.12& 10.23&  15.45&  14.99&  8.48& 9.26& 9.18& 7.97& 8.53& 8.54& 7.78& 8.86& 8.92\\
8&  61.14&  44.80& 33.28& 13.47&  17.67&  17.67 &13.4&  14.88&  14.86&  12.64&  12.68&  12.68&  12.13&  12.98 &12.98
\\\Xhline{5\arrayrulewidth}
\end{tabular}
\label{table:registration}
\end{table*}

\subsection{Registration-oriented Downsampling}
\label{subsec:reg_exp}

Registration aims to  predict rigid transformations between two point clouds, including a rotation and a translation, which can well align them. 
As an overdetermined problem, only a few key points, which can well capture the shape information of a point cloud, are usually extracted, and the registration will be conducted on the key points rather than original point clouds to save memory and computational complexity.  Thus, the registration accuracy also depends on the quality of selected key points. 
In this section, we examined the performance of MOPS-Net with registration as the subsequent task. 

We utilized  point clouds with 1024 points each in ModelNet40  as the original data. 
The paired data were generated by applying random rotations and translations 
to training point clouds. We  adopted  PCRNet~\cite{sarode2019pcrnet} with one iteration as the task network, whose loss function is the L2 difference between the predicted quaternions and ground-truth ones. We quantitatively evaluated different downsampling methods by using the mean rotation error (MRE) between the predicted rotations and ground-truth ones. Note that the MRE of PCRNet trained and tested with original point clouds is 7.21. For MOPS-Net, we set $\tau_{min}=0.1, \alpha=1$ and initialized the learning rate to $1e^{-4}$ and exponentially decreased to $1e^{-5}$ within 250 epochs.

\begin{figure*}[htp!]
     \centering
         \includegraphics[width=\textwidth]{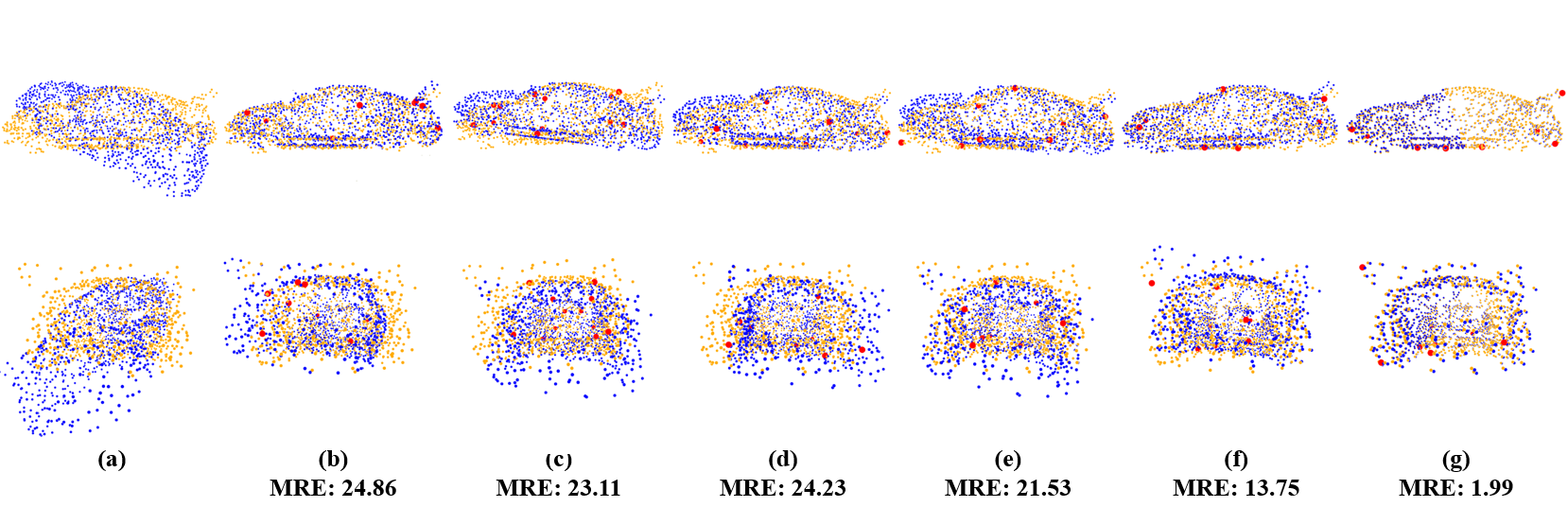}
        \caption{Visual comparisons of registered point clouds by different downsampling methods ($m=8$). (a) Non-registered input pair;  Registered results on the key points extracted by (b) Random sampling; (c) Voxelization; (d) FPS; (e) S-Net; (f)  SampleNet; and (g) MOPS-Net.}
        \label{fig:reg}
\end{figure*}

\textbf{Quantitative and visual comparisons}.
Table~\ref{table:registration} lists MRE values of different downsampling methods with the task network fixed to the pre-trained PCRNet,  
where we can observe that the proposed MOPS-Net can achieve best performance than the traditional methods, S-Net, and SampleNet for all settings, and FMOPS-Net even achieves better performance than MOPS-Net over the generated sets. 
Besides, S-Net suffers from significant performance degradation once the generated point sets are restricted to be subsets of original point clouds (i.e. the matched sets). 
Figure~\ref{fig:reg} visually compares different methods, where the advantage of our MOPS-Net is verified again. 

\textbf{Joint training}. 
In all the above experiments for classification, reconstruction, and registration, the task networks were fixed to be the pre-trained models. 
Here, taking the registration task as an example, we illustrated the advantage of joint training 
i.e., the task network PCRNet and the downsampling network MOPS-Net are jointly trained.  
As listed in Table~\ref{table:joint}, such a joint training manner can further improve the registration accuracy.

\begin{table}[t]
\centering
\caption{Quantitative comparisons of MREs for pre-trained and  jointly trained registration networks. The smaller, the better.}\vspace{-0.1in}
\begin{tabular}{c||ccc|ccc}\Xhline{5\arrayrulewidth}
 &\multicolumn{3}{c|}{Pre-trained PCRNet} & \multicolumn{3}{c}{Jointly trained PCRNet} \\
$m$ & \makebox[3em]{G}&\makebox[3em]{M}&\makebox[3em]{C}& \makebox[3em]{G}&\makebox[3em]{M}&\makebox[3em]{C} \\\Xhline{2\arrayrulewidth}
64& 8.58 & 8.56 & 8.00& 5.96& 8.18& 8.22 \\
32& 7.97 & 8.53 & 8.54& 6.74& 9.18& 9.65 \\
8 & 12.63 & 12.68 &12.68& 6.94& 11.15&  11.15
\\\Xhline{5\arrayrulewidth}
\end{tabular}
\label{table:joint}
\end{table}

\subsection{Robustness Analysis}
\label{subsec:robust_exp}
\begin{figure}[t]
     \centering
         \centering
         \subfigure[]{\includegraphics[width=1.7in]{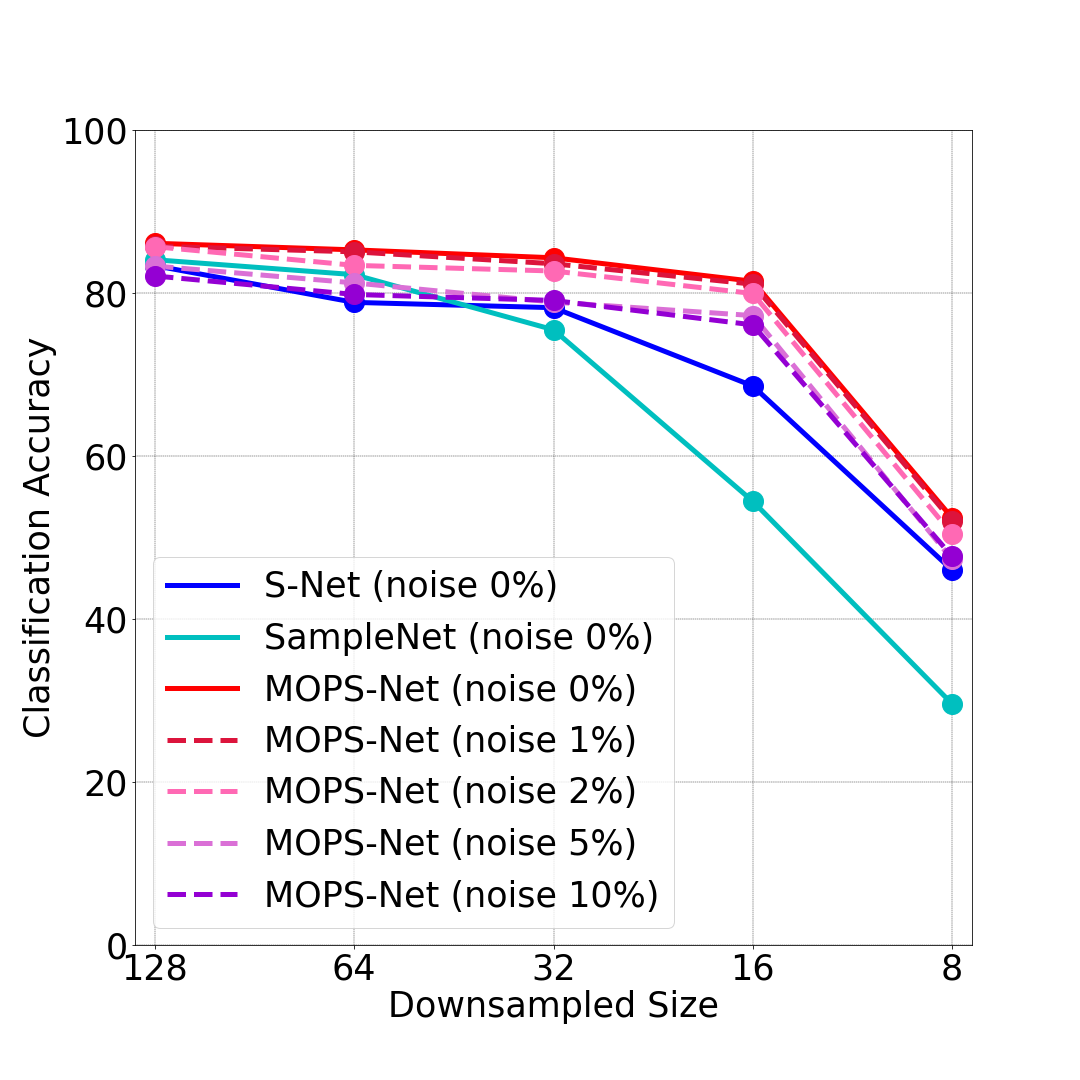}}
         \subfigure[]{\includegraphics[width=1.7in]{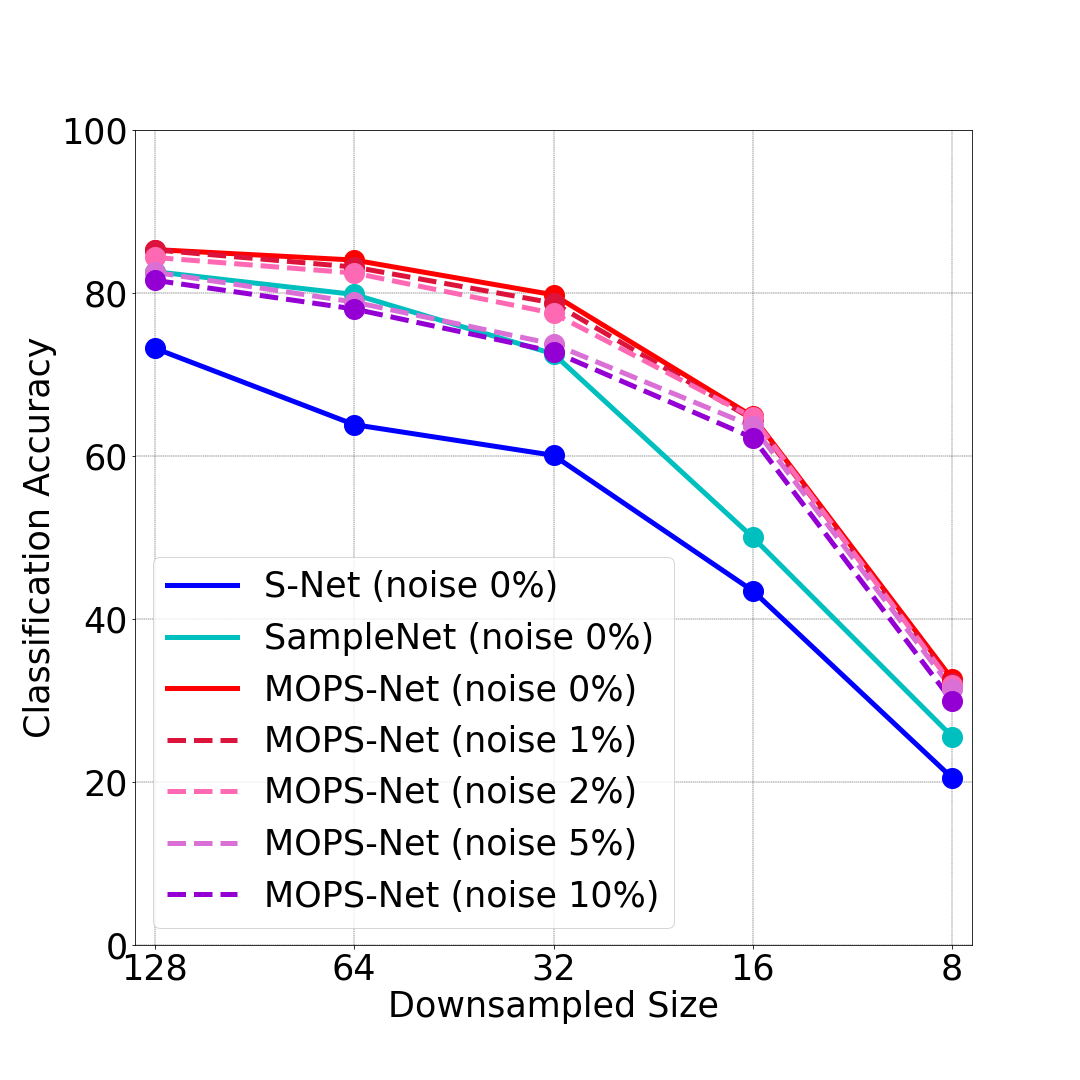}} \vspace{-0.4cm}
        \caption{Comparison of the classification performance of different downsampling methods applied to  point clouds with various levels of noise. (a) performance on the generated sets 
        (b) performance on the completed sets 
        Note that the noise level refers to that added to each dimensional of 3D point cloud data.
        }
        \label{fig:noise_chart}
\end{figure}

We also evaluated the robustness of the proposed MOPS-Net to noise over the classification task. 
We added various levels of the Gaussian noise to input point clouds. As shown in Figure~\ref{fig:noise_chart}, 
even the input point clouds are highly contaminated, i.e., the noise level in each dimension is 10\%, MOPS-Net still remains high accuracy which is comparable to that of MOPS-Net with clean input, demonstrating its robustness. Besides, 
the accuracy of MOPS-Net with noisy input is still higher than that of 
S-Net and SampleNet with clean input. 

\begin{figure*}[htp!]
     \centering
         \includegraphics[width=0.9\textwidth]{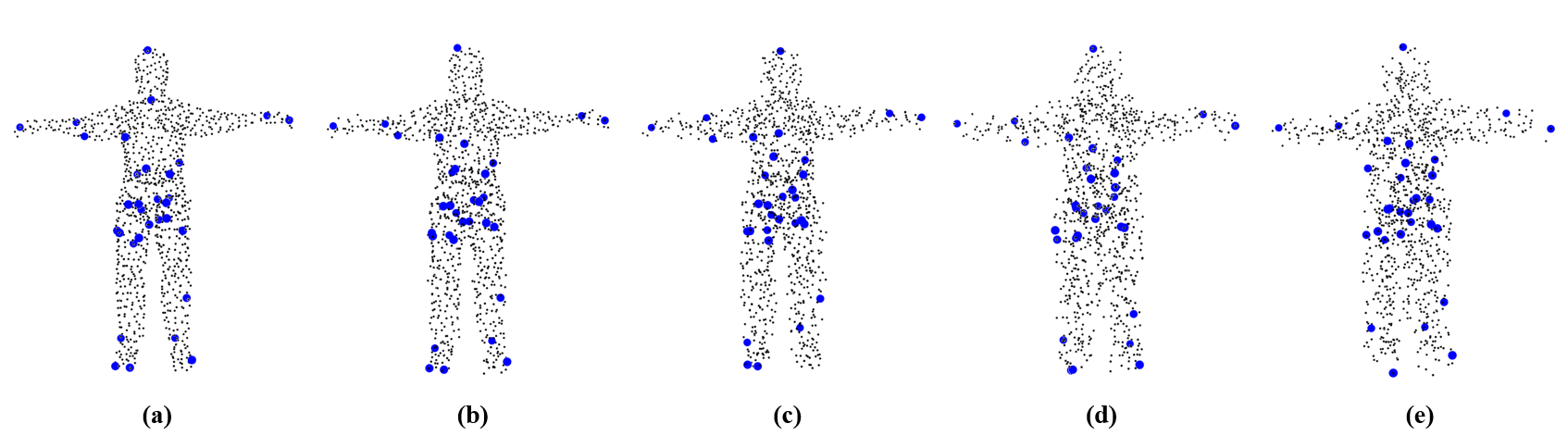}
        \caption{Visualization of sampled point clouds by MOPS-Net on clean data and noisy data with various levels of noise. (a) clean;
        (b) 1\% noise ; (c) 2\% noise; (d) 5\% noise; (e) 10\% noise. Note that the noise level refers to that added to each dimensional of 3D point clouds.}
        \label{fig:noise_visual}
\end{figure*}

In Figure~\ref{fig:noise_visual}, we visually illustrated the downsampled points by our MOPS-Net over noisy data, 
where it can be seen that the proposed MOPS-Net can capture the important regions (head, hands, legs) and the locations of sampled points remain consistent at different noise levels. 


\subsection{Ablation Study}
\label{subsec:ablation}

Taking the reconstruction task in Section \ref{subsubsec:exp_recsmall} as an example, we investigated the effect of the hyper-parameters contained in the loss functions of our MOPS-Net and SampleNet~\cite{lang2020samplenet}. The loss function used by SampleNet is given as 
\begin{equation}
    L_{SampleNet} = L_{task} + \beta*L_{projection} + \gamma*L_{simplify},
\end{equation}
which contains two hyper-parameters $\beta$ and $\gamma$.
Note that in the experiments of Section \ref{subsubsec:exp_recsmall}, the hyper-parameters of both MOPS-Net and SampleNet have been tuned to be almost optimal, i.e.,    $\tau_{min}=0.5$ and  $\alpha=0.2$ for MOPS-Net and $\beta=1e^{-4}$ and $\gamma=5e^{-5}$ for SampleNet. 

\begin{figure}[htp!]
     \subfigure[SampleNet]{\includegraphics[width=1.7in]{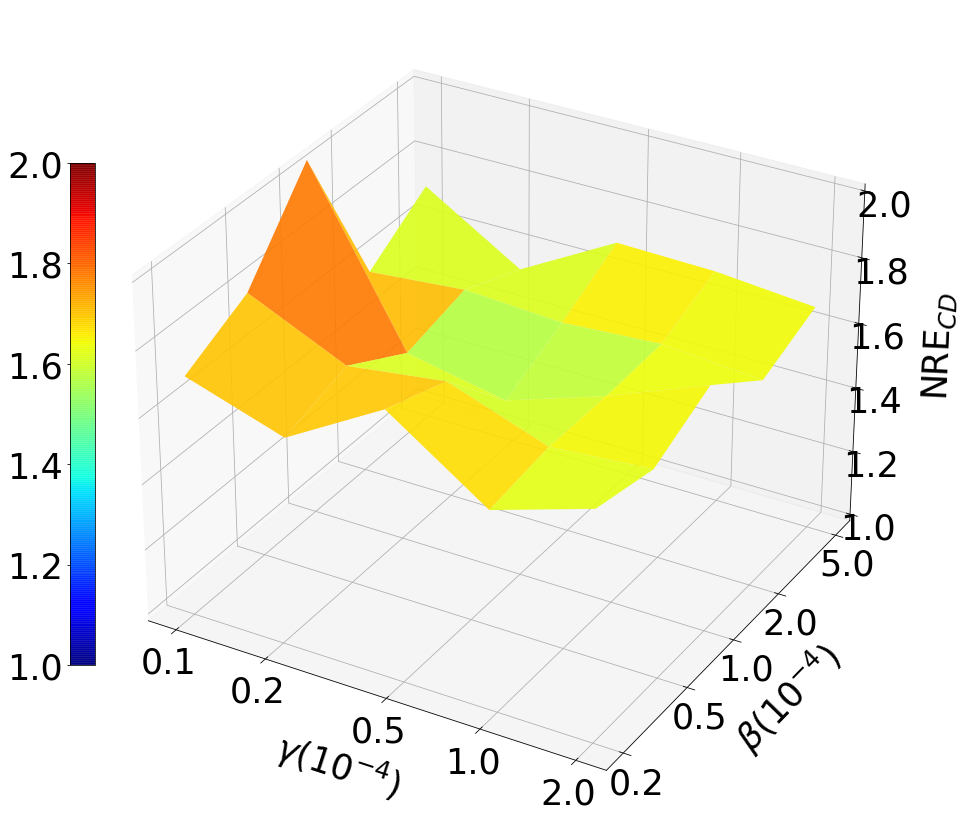}
     \includegraphics[width=1.7in]{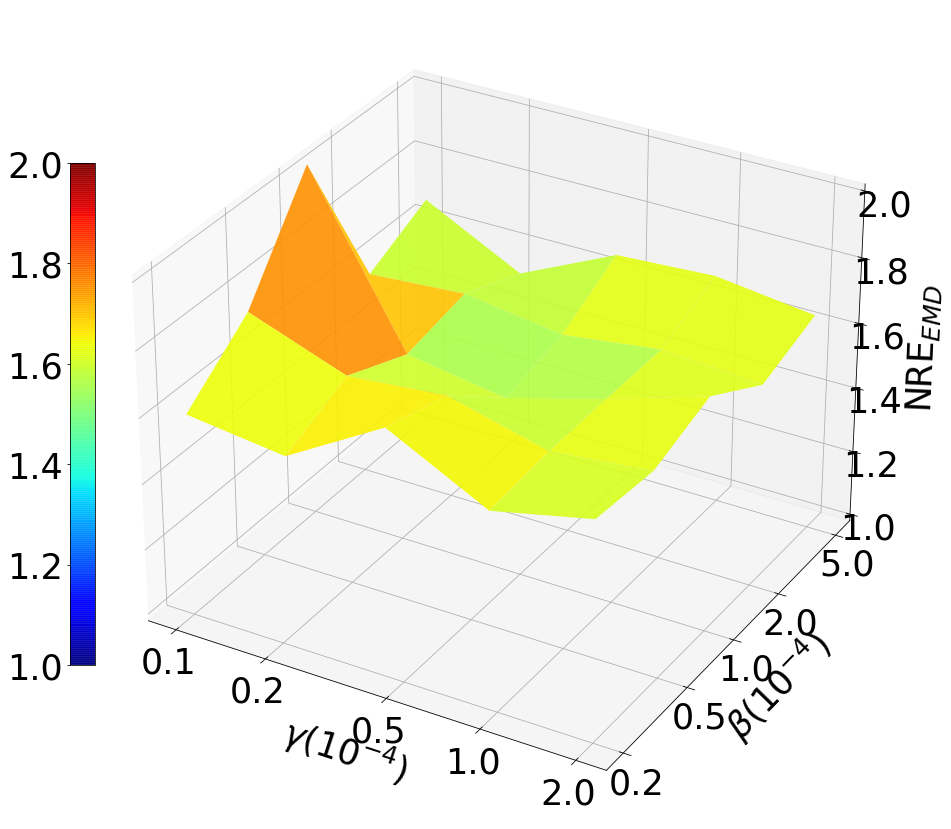}}
         \subfigure[MOPS-Net]{\includegraphics[width=1.7in]{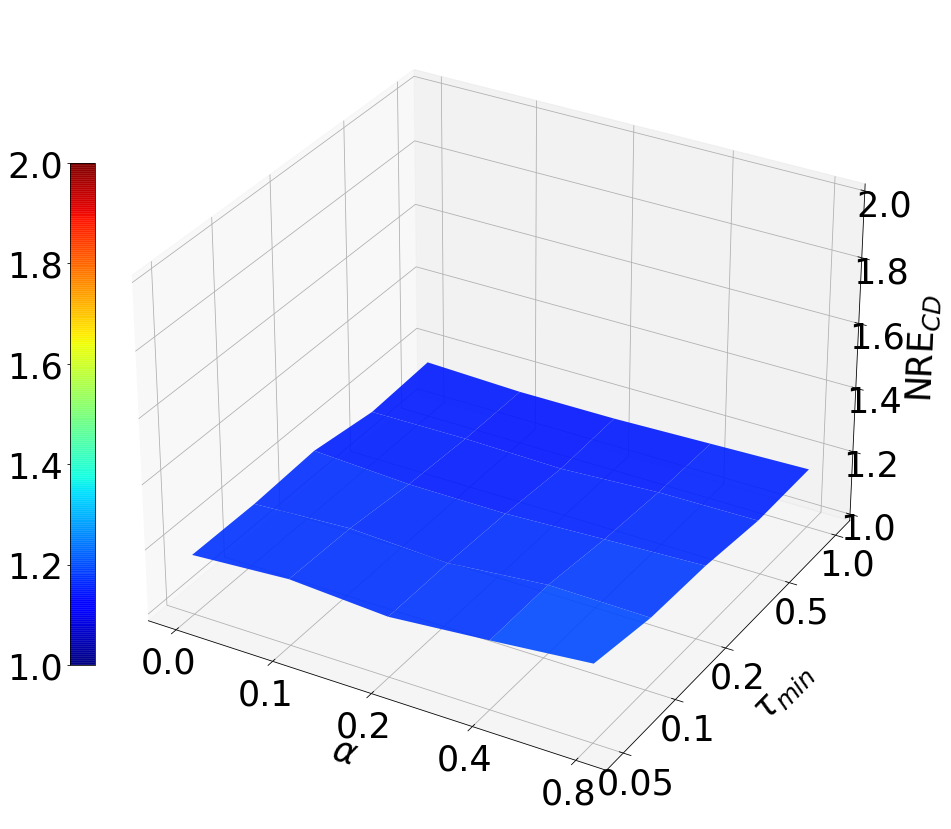}
         \includegraphics[width=1.7in]{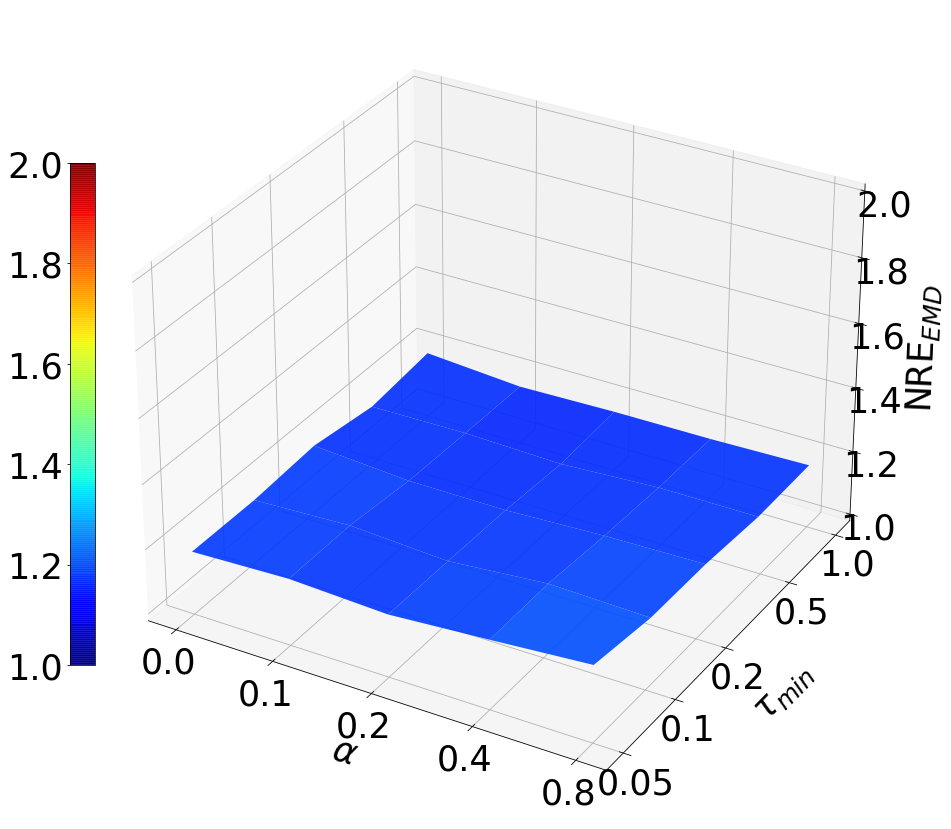}}\vspace{-0.4cm}
        \caption{Illustration of the effect of the hyper-parameters involved in the loss functions of our MOPS-Net and SampleNet on reconstruction performance ($m=32$). SampleNet's performance is highly sensitive to the hyper-parameters, whereas ours is not. 
        }
        \label{fig:ablation}
\end{figure}

As shown in Figure~\ref{fig:ablation}, 
we can see that our MOPS-Net can consistently achieve almost optimal performance in wide ranges of $\alpha$ and $\tau_{min}$, 
demonstrating its stability. However, the performance of SampleNet varies largely as the values of $\beta$ and $\gamma$ change. Besides, under the best parameter settings, SampleNet is still worse than MOPS-Net in terms of both $\textsf{NRE}_{\textsf{CD}}$ and $\textsf{NRE}_{\textsf{EMD}}$.

\subsection{Complexity Analysis}
\label{subsec:complexity}

Table~\ref{table:efficiency_1024} reports the running time of various methods applied to downsample point clouds with 1024 points each. 
All methods were implemented on GTX 2080Ti GPU, and we reported the average inference time per shape. 
From Table~\ref{table:efficiency_1024}, we observe that the running time of FPS is linearly proportional to the downsampled size $m$, 
while the other methods are insensitive to $m$. 
Besides, S-Net and MOPS-Net are even faster than random sampling. 
Compared with S-Net, SampleNet requires an additional projection operation involving $k$-NN search, and thus takes more time than S-Net. 

Besides, we also analyzed the complexity of downsampling large-scale point clouds. The running time and memory consumptions were recorded during the inference period on Quadro RTX 8000 GPU. Table~\ref{table:efficiency_40k} lists the running time of our MOPS-Net applied to downsample point clouds with 40,960 points each. We also provided the running time of random sampling and FPS as a reference. We observe that our MOPS-Net works efficiently and is faster than FPS. Table~\ref{table:memory} lists the memory consumption and running time for each step 
when downsampling 4096 points from 40,960 points, where it can be observed that performing the standard feature extraction via PointNet on large-scale point clouds consumes most of the inference time. By contrast, the proposed sampling modules, including the learning of the sampling matrix and regression of the sampled set are very efficient. Besides, it is worth noticing that the MLP operators which formulate the feature extraction module and learning of the sampling matrix module, requires large memory consumption when dealing with large-scale point clouds.



\begin{table}[htp!]
\centering
\caption{Average running time ($\times 10^{-4}$ seconds) for downsampling a 1024-point model.\textcolor{red}{}}\vspace{-0.1in}
\begin{tabular}{c||ccccc}\Xhline{5\arrayrulewidth}
$m$ & \makebox[3em]{RS}  & \makebox[3em]{FPS}  & \makebox[5em]{S-Net \cite{dovrat2019learning}}  & \makebox[5em]{SampleNet \cite{lang2020samplenet}}  & \makebox[4em]{MOPS-Net}\\\Xhline{2\arrayrulewidth}
512 & 1.22&   68.28&  0.78&   3.39&   1.00 \\
256&  1.26&   29.02&  0.77&   2.95&   0.95\\
128&  1.22&   15.32&  0.75&   2.72&   1.06\\
64&   1.27&   8.68&   0.79&   2.50&   0.99\\
32&   1.32&   5.12&   0.79&   2.39&   0.89\\
16&   1.12&   2.47&   0.72&   2.38&   0.87\\
8&  1.16&   1.46&   0.73&   2.33&   0.98
\\\Xhline{5\arrayrulewidth}
\end{tabular}
\label{table:efficiency_1024}
\end{table}


\begin{table}[htp!]
\centering
\caption{Time efficiency ($\times 10^{-1}$ seconds) for downsampling 40,960 points.}\vspace{-0.1in}
\begin{tabular}{c||ccc}\Xhline{5\arrayrulewidth}
$m$ & \makebox[5em]{RS}  & \makebox[5em]{FPS}  & \makebox[5em]{MOPS-Net}\\\Xhline{2\arrayrulewidth}
4096& 0.07& 182.77& 7.24\\
1024& 0.07& 65.46&  7.04\\
256&  0.07& 16.31&  7.40\\
\Xhline{5\arrayrulewidth}
\end{tabular}
\label{table:efficiency_40k}
\end{table}

\begin{table}[t]
\centering
\caption{Memory consumption and running time of MOPS-Net to downsample $m=4096$ points from large-scale point clouds with $n=40,960$ points.}\vspace{-0.1in}
\begin{tabular}{c||c|c}\Xhline{5\arrayrulewidth}
 Module & GPU Memory &  Time ($\times 10^{-1}$)\\\Xhline{2\arrayrulewidth}
Feature extraction & 2123MB & 6.59 \\
Learn sampling & 2560MB & 0.04\\
Regress points  & 0MB & 0.14\\
Task network & 4MB & 0.25\\
\Xhline{5\arrayrulewidth}
\end{tabular}
\label{table:memory}
\end{table}

\section{Conclusion \& Future Work}
\label{sec:con}
In this paper, we presented MOPS-Net, a novel end-to-end deep learning framework for task-oriented point cloud downsampling. 
In contrast to the existing methods, we designed MOPS-Net from the perspective of matrix optimization. As the original discrete and combinatorial optimization problem is difficult to solve, we obtained a continuous and differentiable form by relaxing the 0-1 constraint of each variable. MOPS-Net elegantly mimics the function of the resulting matrix optimization problem by exploring both local and global structures of input data. MOPS-Net is permutation invariant and can be end-to-end trained with a task network. We applied  MOPS-Net to three typical applications,  including 3D point cloud classification, reconstruction, and registration, and observed that MOPS-Net produces better results than state-of-the-art methods. Moreover, MOPS-Net is flexible in that with simple modification, a single network with one-time training can handle arbitrary downsampling ratios. We justified our optimization driven design principle and demonstrated the efficacy of MPOS-Net through extensive evaluations and comparisons. 

The promising results of MOPS-Net inspire several interesting future directions. For example, it can replace the widely used FPS in feature extraction of current networks to boost performance. Though MOPS-Net is designed for point cloud downsampling, increasing the dimension of differential sampling matrix allows us to handle upsampling as well. Moreover, MOPS-Net opens the door to apply matrix optimization in deep learning. We believe the matrix optimization idea is general and can work for other selection and ranking problems, such as key frame selection in videos~\cite{wu2019adaframe,mei2020patch}, band selection in hyperspectral images~\cite{guo2006band,wang2016salient} and view selection in light field images~\cite{jin2020deep}.

\balance

\end{document}